\documentclass{article}

\usepackage[preprint]{neurips_2026}
\makeatletter
\def\@noticestring{\textsuperscript{*}\,Equal contribution.}
\makeatother
\usepackage{amsmath}
\usepackage{tcolorbox}
\usepackage{xcolor}
\newcommand{\delt}[1]{{\color{gray}\small #1}}

\usepackage[utf8]{inputenc} 
\usepackage[T1]{fontenc}    
\usepackage{hyperref}       
\usepackage{url}            
\usepackage{booktabs}       
\usepackage{amsfonts}       
\usepackage{nicefrac}       
\usepackage{microtype}      
\usepackage{xcolor}         
\usepackage{graphicx}
\usepackage{multirow}
\usepackage{array}
\usepackage{soul}
\usepackage{cleveref}
\usepackage{wrapfig}

\title{Post-Reasoning: Improving the Performance of Non-Thinking Models at Negligible Cost}

\author{%
  \textbf{Richmond Sin Jing Xuan}\textsuperscript{1,*} \quad 
  \textbf{Rishabh Bhardwaj}\textsuperscript{1,*} \\ 
  \textbf{Soujanya Poria}\textsuperscript{1}\\
  \textsuperscript{1}Nanyang Technological University \\ 
  \texttt{e250239@e.ntu.edu.sg, \{rishabh.bhardwaj, soujanya.poria\}@ntu.edu.sg}
}

\begin{document}

\maketitle

\begin{abstract}
As the widespread adoption of Large Language Models (LLMs) accelerates, token consumption from intermediate reasoning traces increasingly contributes to inference latency and operational cost. Recent studies suggest that many real-world tasks require little to no explicit reasoning, with additional reasoning sometimes even degrading performance. In this work, we propose \textbf{Post-Reasoning}, a simple yet effective approach that improves instruction-tuned models by conditioning them to justify their answers after generating the final response. By design, it enables the final answer to be obtained with no reasoning tokens before the answer and with negligible latency and token cost, while still improving performance through simple instruction augmentation. We evaluate Post-Reasoning across \(117\) model--benchmark settings spanning \(13\) open and proprietary models, \(7\) model families, and \(9\) diverse reasoning and knowledge-intensive benchmarks, including AMC, HMMT, GSM8K, GPQA, MMLU-Pro, and BIG-Bench Hard. Post-Reasoning improves performance in \(84.62\%\) of evaluated settings, achieving a mean relative improvement of \(15.57\%\). Furthermore, we propose supervised post-reason tuning, which further improves performance in \(91.11\%\) of evaluated settings, and exceeds the prompt-based post-reasoning baseline by an average of \(7.17\%\), demonstrating that post-reasoning can be effectively internalized through training. Ultimately, Post-Reasoning establishes a new performance ceiling for direct-answer capabilities.
\end{abstract}

\section{Introduction}
As Large Language Model (LLM) adoption accelerates, with over a billion people using AI in some capacity~\citep{microsoft2025globalai}, token consumption has surged~\citep{aubakirova2025stateofai}, making efficient inference a critical operational challenge~\citep{openai2025enterprise}. Beyond the growing user base, token consumption is further amplified by reasoning traces (intermediate tokens) generated prior to producing final responses~\citep{deloitte2026aiinfra}. Consequently, maximizing the number of queries served under a fixed token budget is emerging as a central optimization problem for both LLM-based system providers (e.g., Anthropic, OpenAI, Lovable) and consumers.

In reasoning-enabled models, a substantial portion of the token budget is consumed by reasoning traces rather than the final answer, leading to increased inference cost and latency per task~\citep{sui2025stop, chen2024not, luo2025o1, yeo2025demystifying, han2024token, ma2025cot, hao2024training}. Such models can consume on the order of 10$\times$ more tokens without guaranteed performance gains~\citep{srivastava2026llms}. Real-world tasks for which LLMs are used lie on a spectrum of token consumption, where logic-intensive tasks such as advanced mathematics (e.g., MATH, AIME) and complex coding benchmarks (e.g., HumanEval, Codeforces) benefit from extended reasoning and occupy the higher end of the spectrum. In contrast, simpler benchmarks (e.g., GSM8K) and tasks such as summarization, factual question answering, and basic arithmetic often require little to no reasoning, with reduced reasoning sometimes matching or outperforming full reasoning~\citep{li2025thinkless,aggarwal2025optimalthinkingbench}, placing them toward the lower end; Appendix~\ref{app:when_to_think} provides a practical decision rule based on task complexity and the available latency and token budget.

In this work, we propose a novel approach, \textbf{Post-Reasoning}, that improves the performance of instruction-tuned models, i.e., models that directly generate responses without explicit reasoning traces and therefore lie on the extreme left of the token spectrum. Post-Reasoning conditions a model to generate additional tokens after producing the final answer and, unlike pre-reasoning approaches, generates no reasoning tokens before the answer, so its added latency and token cost to the answer are negligible (Appendix~\ref{app:latency_results}). Post-reasoning behavior can be induced in two ways: (1) through prompting, by instructing the model to first provide the answer and then justify it (e.g., ``What is \(10+2\)? State the final answer immediately and justify your answer.''), and (2) through supervised post-reason tuning, where the behavior is internalized into the model weights. Furthermore, the post-reasoning generation can optionally be omitted at inference time through stop-token-based decoding, allowing the final answer to be obtained without generating the additional justification tokens.

\begin{wrapfigure}{r}{0.5\textwidth}
    \centering
    \vspace{-10pt}
    \includegraphics[width=0.5\textwidth]{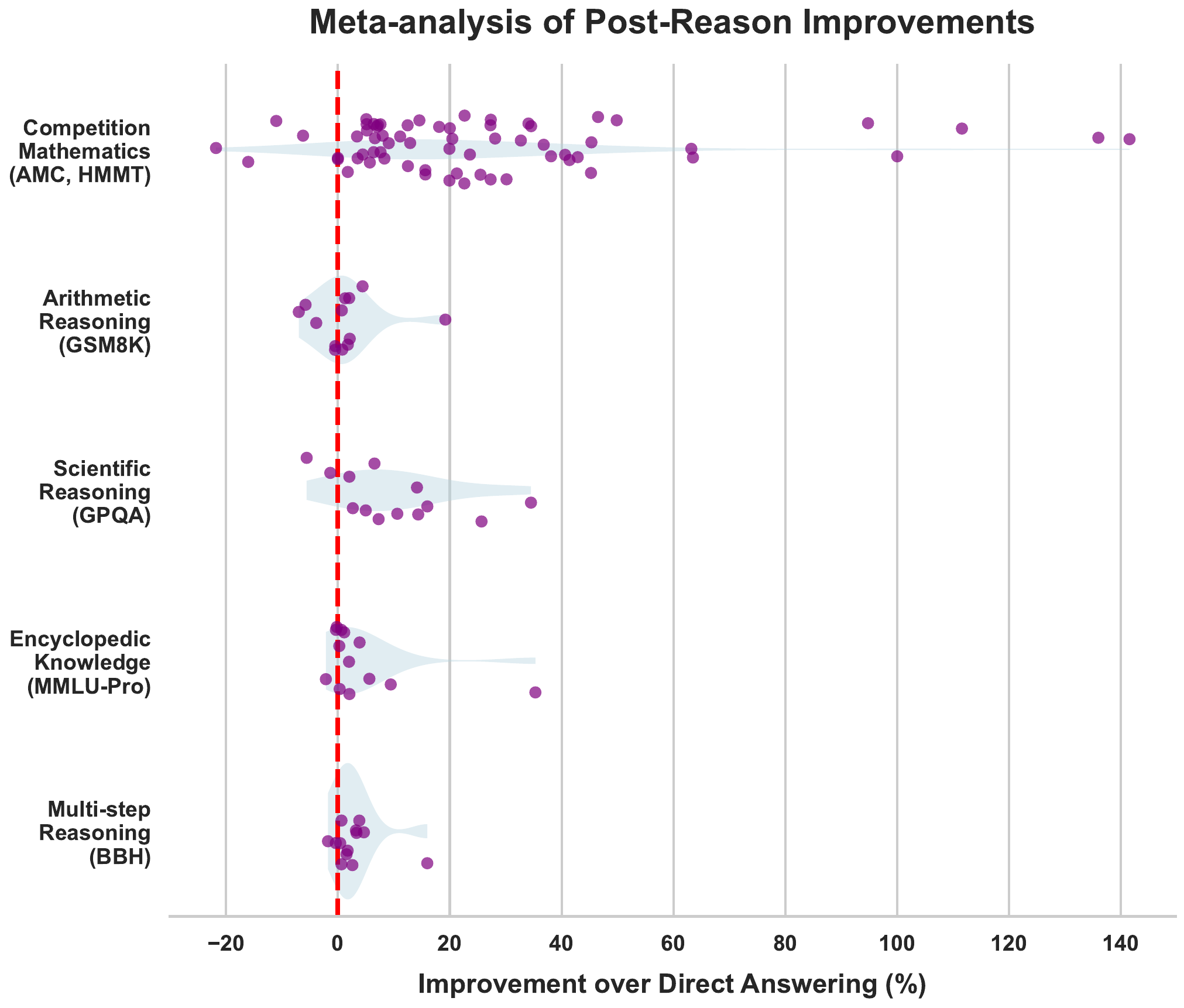}
    \caption{{Meta-analysis of Post-Reasoning improvements across benchmarks. Each point denotes the relative gain over direct answering for a (model, task) pair. Gains are predominantly positive, with larger improvements on multi-step reasoning tasks (AMC, HMMT, MATH) and smaller or mixed gains on arithmetic and knowledge-intensive benchmarks.}}
    \label{fig:intro-figure}
    \vspace{-5pt}
\end{wrapfigure}

Experiments were conducted across 13 open-source and proprietary models spanning different scales (4B--70B) and model families (Llama, Gemma, Mistral, Qwen, Gemini, Claude, GPT). Across all evaluations \Cref{fig:intro-figure}, Post-Reasoning improves model performance in 84.62\% of settings, yielding average gains of 24.99\% on competition mathematics (AMC 8/10/12), 23.11\% on olympiad-style reasoning (HMMT), 1.13\% on standard arithmetic reasoning (GSM8K), 10.14\% on graduate-level scientific reasoning (GPQA Main), 4.45\% on multi-domain knowledge reasoning (MMLU-Pro), and 3.16\% on multi-step logical reasoning (BBH). Furthermore, supervised post-reason tuning on 10 open-source models demonstrates strong generalization, improving performance in 91.11\% of evaluated settings and achieving significant gains across benchmarks (44.49\% on AMC, 41.97\% on HMMT, 2.80\% on GSM8K, 12.65\% on GPQA Main, 4.26\% on MMLU-Pro, and 2.16\% on BBH). Notably, post-reason tuned models outperform prompt-based Post-Reasoning in 73.33\% of evaluated settings, suggesting that post-reasoning behavior can be effectively internalized through supervised training. In summary, our core contributions are as follows:

\begin{itemize}
    \item \textbf{Domain-generic Improvement on Instruct Models}: We introduce Post-Reasoning, a novel approach that improves the performance of instruction-tuned models with negligible additional inference latency and token cost for obtaining the final answer.
    
    \item \textbf{Post-Reason Prompting}: We demonstrate that conditioning a model to post-reason consistently improves direct-answer accuracy across both open and proprietary models. Across 13 models spanning 4B--70B parameters and multiple benchmark suites, post-reasoning improves performance in 84.62\% of evaluated settings, with gains exceeding 100\% on several long-horizon reasoning tasks.
    
    \item \textbf{Supervised Post-Reason Tuning}: We propose a supervised post-reason tuning framework that optimizes models exclusively on post-answer justifications using masked loss optimization. Post-reason tuning further improves performance over prompt-based post-reasoning in 73.33\% of evaluated model--task combinations, while enabling strong cross-domain generalization beyond the training distribution.
\end{itemize}

\paragraph{Related Work.} Chain-of-Thought (CoT) prompting~\citep{wei2022chain} and subsequent reasoning-based approaches~\citep{wang2023self,yao2023tree,lightman2023let,openai2025gptoss120bgptoss20bmodel,deepseekai2025deepseekr1incentivizingreasoningcapability} have demonstrated that generating intermediate reasoning traces can substantially improve performance on complex reasoning tasks. However, these gains often come at the cost of significantly increased inference-time token consumption, latency, and operational expense. Recent works therefore explore reasoning compression, pruning, and selective thinking strategies~\citep{li2025thinkless,aggarwal2025optimalthinkingbench,srivastava2026llms}, showing that many tasks require little or no explicit reasoning and that excessive reasoning can sometimes degrade performance.

Our work is orthogonal to these approaches. Rather than reducing pre-answer reasoning traces, we investigate whether reasoning generated \emph{after} the answer can still improve performance while avoiding additional latency before the final response. Furthermore, unlike prior reasoning SFT approaches that jointly train on reasoning traces and answers, our post-reason tuning framework supervises models only on post-answer justifications, enabling reasoning improvements without modifying the direct-answer generation process and yielding stronger cross-domain generalization beyond the training distribution. Closest to our work, answer-first distillation trains small students on teacher rationales placed after the label~\citep{chen2024pst,wadhwa2024mysteries}; answer-first fine-tuning underperforms no reasoning on vision tasks~\citep{li2025thinkafteranswer,wang2025auxthink}; and TwT distills reasoning out of the output~\citep{xu2025twtthinkingtokenshabitual}. We show that an instruction alone improves LLMs without training, and we train on the model's own justifications with no answer loss, unlike~\citet{hsieh2023distilling}.

\section{Methodology}

Let $\mathcal{V}$ denote the tokenizer vocabulary. Given an input prompt $x \in \mathcal{V}^D$, the model generates a response $y \in \mathcal{V}$ (for simplicity, treated as a single token). Let $\mathrm{LM}_\theta$ denote a language model parameterized by $\theta$, which maps an input sequence to a distribution over $\mathcal{V}$. In reasoning mode, the output $y$ is conditioned not only on the input $x$, but also on two additional factors: (\textbf{F1}) an augmented instruction $\delta_t$ that prompts the model to reason (e.g., ``What is $2+3+10$? \textit{Think step by step}''), and (\textbf{F2}) auto-regressively generated reasoning tokens $c_t$ (e.g., ``$2+3=5$, $5+10=15$''), which are produced before the final answer. Formally,
\begin{equation}
c_t \sim \mathrm{LM}_\theta(\cdot \mid x, \delta_t), \quad
y \sim \mathrm{LM}_\theta(\cdot \mid x, \delta_t, c_t).
\end{equation}

Here, the second factor, i.e., $c_t$, is the primary contributor to increased latency and cost in obtaining the LLM response. In \textbf{Post-Reasoning}, we omit this factor from conditioning the answer and instead leverage the augmented instruction (F1) to help condition the response. To this end, we instruct the model to ``state the final result first, then justify,'' denoted by $\delta_p$:
\begin{equation}
y \sim \mathrm{LM}_\theta(\cdot \mid x, \delta_p), \quad
c_p \sim \mathrm{LM}_\theta(\cdot \mid x, \delta_p, y).
\end{equation}

Here, $c_p$ denotes post-answer (post-justification) tokens. These tokens do not contribute to the response cost when generation is truncated immediately after producing the final answer $y$, for example by specifying a designated stop token (or end-of-answer marker) and terminating autoregressive decoding once this token is generated.\footnote{Optionally, generation can be continued beyond this point to produce $c_p$, providing interpretable justifications for the predicted outcome.}

\subsection{Prompt-based Post-Reasoning}
Similar to Chain-of-Thought prompting which asks model to first think then answer, we instruct models to generate justifications for their answers. The exact prompt templates and instruction formats are provided in Appendix \ref{app:prompts_all}.

\begin{tcolorbox}[colback=gray!5,colframe=black!70,title=Post-Reason Prompt]
{\raggedright\texttt{<task-specific instruction> State the final answer immediately as "Answer:\ [X]", then justify it.}\par}
\end{tcolorbox}

\subsection{Supervised Post-Reason Tuning} \label{sec:loss}
To train the model for post-reasoning, we optimize the likelihood of post-reasoning tokens while masking the loss on the answer tokens. Specifically, given an input $x$, a target answer $y$, and post-reasoning tokens $c_p = (c_1, \dots, c_T)$, the model is trained to predict the justification sequence conditioned on both the input and the answer:
\begin{equation}
\mathcal{L} = -\sum_{i=1}^{T} \log P_{\theta}(c_i \mid x, y, c_{<i})
\label{eq:loss_function}
\end{equation}
By restricting supervision to the post-reasoning tokens, the model is encouraged to generate coherent justifications conditioned on the answer, rather than overfitting to answer prediction itself.

\paragraph{Intuition behind post-reasoning.} In autoregressive next-token prediction, each generated token is conditioned on the preceding context. As discussed earlier, Chain-of-Thought (CoT) prompting introduces two forms of conditioning: \(\mathbf{F1}\), the augmented instruction (e.g., ``Think step by step''), and \(\mathbf{F2}\), the generated reasoning tokens themselves\citep{wei2022chain}. While prior work primarily attributes CoT improvements to the reasoning traces (\(\mathbf{F2}\)), the impact of the augmented instruction (\(\mathbf{F1}\)) remains underexplored. Post-Reasoning primarily leverages this instruction-level conditioning, suggesting that performance gains may arise not only from explicit reasoning traces, but also from how models are prompted (conditioned) to structure generation.

\section{Experimental Setup}
\paragraph{Models.} To extensively study the impact of post-reasoning, we consider a range of open-source models across providers and sizes: \(4\)B to \(70\)B parameters: Qwen\(3.5\) (\(4\)B, \(9\)B, \(27\)B) \cite{yang2025qwen3technicalreport}, Llama-\(3.1\) (\(8\)B) and Llama-\(3.3\) (\(70\)B) \cite{grattafiori2024llama3herdmodels}, Gemma-\(3\) (\(12\)B, \(27\)B) \cite{gemmateam2025gemma3technicalreport}, Ministral-\(3\) (\(8\)B, \(14\)B) \cite{liu2026ministral3}, and Mistral-Small (\(24\)B). Since the primary aim is to improve the performance of instruction-tuned models at negligible cost and latency, all the models chosen are either instruction-tuned or facilitate a turn-off thinking mode. Generations were constrained to a maximum of \(2,048\) tokens for instruct and \(4,096\) for Post-Reason models. We set temperature to \(0.0\), top-p to \(0.8\), and top-k to \(20\). 

\paragraph{Benchmarks.}
We evaluate our approach on a diverse suite of benchmarks spanning multiple domains and levels of reasoning complexity. For standard mathematical reasoning, we use GSM8K~\citep{cobbe2021trainingverifierssolvemath}. To assess performance on competition-level and structured problem-solving tasks, we evaluate on AMC (8/10/12) and Harvard-MIT Mathematics Tournament (HMMT) benchmarks \citep{ding2024easy2hard}. To further evaluate broader scientific, knowledge-intensive, and multi-step reasoning capabilities, we include GPQA~\citep{rein2023gpqagraduatelevelgoogleproofqa}, MMLU-Pro, and BIG-Bench Hard (BBH). Collectively, this benchmark suite spans a broad spectrum of task difficulty, ranging from standard arithmetic reasoning to complex scientific, logical, and long-horizon reasoning tasks.

\paragraph{Baseline.} We compare post-reasoning against standard instruction-tuned models operating without reasoning. These baseline results are denoted as \textbf{Direct} in our experiments, while post-reason prompting results are denoted as \textbf{Post}. Notably, we do not directly compare against native pre-reasoning (CoT) models as the primary goal of Post-Reasoning is to improve the performance of standard instruction-tuned models with negligible additional inference latency and token cost. We nevertheless showed that Post-Reason complements native thinking on open-source models and low-effort thinking on proprietary models where token budget is strict in Appendix~\ref{app:thinking_results}; investigating this interaction is beyond the scope of the present study.

\begin{table}[htbp]
\centering
\caption{AMC Competition Mathematics. Direct: standard prompting. Post: post-reason prompting. $\Delta$: Post minus Direct, in accuracy points.}
\label{tab:phase1_amc}
\resizebox{0.8\textwidth}{!}{%
\begin{tabular}{lccccccccc}
\toprule
& \multicolumn{3}{c}{\textbf{AMC 8}} & \multicolumn{3}{c}{\textbf{AMC 10}} & \multicolumn{3}{c}{\textbf{AMC 12}} \\
\cmidrule(lr){2-4} \cmidrule(lr){5-7} \cmidrule(lr){8-10}
\textbf{Model} & Direct & Post & $\Delta$ & Direct & Post & $\Delta$ & Direct & Post & $\Delta$ \\
\midrule

\multicolumn{10}{l}{\textbf{Llama Family}} \\
\quad Llama-\(3.1\) (\(8\)B) 
& 7.09 & \textbf{11.57} & \delt{+4.48}
& 7.42 & \textbf{12.13} & \delt{+4.71}
& 4.43 & \textbf{10.70} & \delt{+6.27} \\

\quad Llama-\(3.3\) (\(70\)B) 
& 19.78 & \textbf{28.73} & \delt{+8.95}
& 27.42 & \textbf{33.03} & \delt{+5.61}
& 20.30 & \textbf{26.94} & \delt{+6.64} \\

\midrule
\multicolumn{10}{l}{\textbf{Gemma Family}} \\
\quad Gemma-\(3\) (\(12\)B) 
& 16.04 & \textbf{17.16} & \delt{+1.12}
& 9.89 & \textbf{13.26} & \delt{+3.37}
& 8.49 & 6.64 & \delt{-1.85} \\

\quad Gemma-\(3\) (\(27\)B) 
& 10.82 & \textbf{15.30} & \delt{+4.48}
& 13.93 & \textbf{17.08} & \delt{+3.15}
& 11.81 & \textbf{15.13} & \delt{+3.32} \\

\midrule
\multicolumn{10}{l}{\textbf{Mistral \& Ministral}} \\
\quad Ministral-\(3\) (\(8\)B) 
& 13.06 & \textbf{18.66} & \delt{+5.60}
& 10.11 & \textbf{11.69} & \delt{+1.58}
& 9.23 & 7.75 & \delt{-1.48} \\

\quad Ministral-\(3\) (\(14\)B) 
& 17.54 & \textbf{21.27} & \delt{+3.73}
& 12.36 & \textbf{15.51} & \delt{+3.15}
& 7.01 & \textbf{9.59} & \delt{+2.58} \\

\quad Mistral-Small (\(24\)B) 
& 7.09 & \textbf{13.81} & \delt{+6.72}
& 12.13 & \textbf{13.48} & \delt{+1.35}
& 6.27 & \textbf{7.75} & \delt{+1.48} \\

\midrule
\multicolumn{10}{l}{\textbf{Qwen Family}} \\
\quad Qwen\(3.5\) (\(4\)B) 
& 16.42 & \textbf{20.90} & \delt{+4.48}
& 8.99 & \textbf{10.79} & \delt{+1.80}
& 4.06 & \textbf{5.17} & \delt{+1.11} \\

\quad Qwen\(3.5\) (\(9\)B) 
& 18.43 & \textbf{21.12} & \delt{+2.69}
& 12.36 & \textbf{16.63} & \delt{+4.27}
& 8.12 & \textbf{9.59} & \delt{+1.47} \\

\quad Qwen\(3.5\) (\(27\)B) 
& 27.76 & \textbf{30.67} & \delt{+2.91}
& 28.76 & \textbf{32.36} & \delt{+3.60}
& 20.30 & \textbf{20.66} & \delt{+0.36} \\

\bottomrule
\end{tabular}%
}
\end{table}

\section{Results and Discussions}

\paragraph{Core Effect of Post-Reasoning.}
Across all 10 models  from 4 families and sizes from 4B to 70B, the Post-Reason baseline outperforms the Direct Answer baseline in \textbf{83.33\%} of evaluated cases, demonstrating that enforcing a post-reasoning sequence structure improves initial token generation without requiring any model updates. 

\subsection{Mathematical Benchmarks.}

\textbf{AMC Benchmarks.}
As shown in \Cref{tab:phase1_amc}, post-reasoning yields consistent improvements across model families, with an average relative gain of \textbf{30.01\%}. When stratified by scale, smaller models ($\leq$10B) achieve larger gains (\textbf{37.71\%}) compared to mid-sized (10--30B, \textbf{23.28\%}) and large models ($\geq$70B, \textbf{32.81\%}), indicating that post-reasoning provides comparatively greater benefits for models with lower baseline accuracy under direct-answer prompting, while remaining effective across all scales.

The magnitude of improvement varies across models and tasks, ranging from modest gains to large increases (e.g., up to \textbf{+8.95 points} on AMC 8 for Llama-3.3-70B), with occasional decreases (e.g., Gemma-3-12B and Ministral-3-8B on AMC 12). Despite this variability, improvements are observed across all AMC difficulty levels (AMC 8/10/12), indicating that post-reasoning acts as a general performance enhancement rather than a task-specific effect.

\textbf{Competition Mathematics (HMMT).}
This trend extends to more challenging competition benchmarks, as shown in \Cref{tab:phase1_math_combined}. On HMMT, post-reasoning improves performance in \textbf{17/20 cases (85\%)}, with the largest gains on HMMT Feb (e.g., +3.13 points for Llama-3.3-70B and +2.41 points for Ministral-3-8B). These results reinforce that post-reasoning is particularly effective for long-horizon, multi-step reasoning tasks.

\textbf{Standard Mathematics (GSM8K).}
In contrast, gains are more modest on GSM8K within the same table, with improvements observed in \textbf{6/10 models (60\%)} and an average relative gain of \textbf{+1.25\%}. This difference reflects task characteristics: AMC and HMMT problems require deeper, multi-step reasoning, whereas GSM8K involves shorter reasoning chains with fewer intermediate steps. Consequently, while post-reasoning remains beneficial, the available headroom is smaller when direct inference is already competitive.

\subsection{General Reasoning Benchmarks.}
We further evaluate post-reasoning beyond mathematical tasks on a diverse set of reasoning and knowledge-intensive benchmarks, including GPQA, MMLU-Pro, and BIG-Bench Hard (\Cref{tab:phase1_ood}). Across these benchmarks, post-reasoning improves performance in \textbf{24/30 cases (80\%)}, demonstrating that its benefits extend across domains.

However, the magnitude of improvement is generally more modest compared to AMC and HMMT, with gains typically in the range of \textbf{0.1--3.2 points}, and occasional larger improvements on specific models (e.g., +15.05 points on GPQA for Qwen-3.5-9B). Improvements are more pronounced on GPQA, which requires deeper reasoning and domain-specific knowledge, whereas gains on MMLU-Pro and BIG-Bench Hard are smaller and more uniform.
As observed earlier, gains vary across models and tasks, suggesting that the effectiveness of post-reasoning depends on both task complexity and model alignment. Altogether, these results indicate that the effectiveness of post-reasoning scales with the depth and complexity of reasoning required by the task, while remaining broadly beneficial across models and domains.

\begin{table}[htbp]
\centering
\caption{Mathematical Reasoning Benchmarks: competition (HMMT) and standard (GSM8K). $\Delta$: Post minus Direct, in accuracy points.}
\label{tab:phase1_math_combined}
\resizebox{0.78\textwidth}{!}{%
\begin{tabular}{lccccccccc}
\toprule
& \multicolumn{3}{c}{\textbf{HMMT Feb}} & \multicolumn{3}{c}{\textbf{HMMT Nov}} & \multicolumn{3}{c}{\textbf{GSM8K}} \\
\cmidrule(lr){2-4} \cmidrule(lr){5-7} \cmidrule(lr){8-10}
\textbf{Model} & Direct & Post & $\Delta$ & Direct & Post & $\Delta$ & Direct & Post & $\Delta$ \\
\midrule

\multicolumn{10}{l}{\textbf{Llama Family}} \\
\quad Llama-\(3.1\) (\(8\)B) 
& 1.44 & \textbf{2.88} & \delt{+1.44}
& 2.45 & 2.18 & \delt{-0.27}
& 11.45 & \textbf{13.65} & \delt{+2.20} \\

\quad Llama-\(3.3\) (\(70\)B) 
& 6.73 & \textbf{9.86} & \delt{+3.13}
& 7.36 & \textbf{10.35} & \delt{+2.99}
& 34.27 & \textbf{35.78} & \delt{+1.51} \\

\midrule
\multicolumn{10}{l}{\textbf{Gemma Family}} \\
\quad Gemma-\(3\) (\(12\)B) 
& 3.61 & \textbf{4.33} & \delt{+0.72}
& 5.18 & \textbf{5.99} & \delt{+0.81}
& 21.30 & \textbf{21.68} & \delt{+0.38} \\

\quad Gemma-\(3\) (\(27\)B) 
& 5.77 & \textbf{6.25} & \delt{+0.48}
& 5.72 & \textbf{7.90} & \delt{+2.18}
& 31.92 & 31.77 & \delt{-0.15} \\

\midrule
\multicolumn{10}{l}{\textbf{Mistral \& Ministral}} \\
\quad Ministral-\(3\) (\(8\)B) 
& 2.16 & \textbf{4.57} & \delt{+2.41}
& 3.00 & \textbf{4.36} & \delt{+1.36}
& 22.74 & 21.15 & \delt{-1.59} \\

\quad Ministral-\(3\) (\(14\)B) 
& 3.85 & \textbf{4.33} & \delt{+0.48}
& 4.90 & \textbf{5.18} & \delt{+0.28}
& 28.89 & 27.22 & \delt{-1.67} \\

\quad Mistral-Small (\(24\)B) 
& 3.85 & 3.61 & \delt{-0.24}
& 6.27 & \textbf{7.08} & \delt{+0.81}
& 29.57 & \textbf{29.95} & \delt{+0.38} \\

\midrule
\multicolumn{10}{l}{\textbf{Qwen Family}} \\
\quad Qwen\(3.5\) (\(4\)B) 
& 3.12 & \textbf{3.37} & \delt{+0.25}
& 2.72 & \textbf{3.54} & \delt{+0.82}
& 19.33 & \textbf{19.48} & \delt{+0.15} \\

\quad Qwen\(3.5\) (\(9\)B) 
& 3.61 & \textbf{4.33} & \delt{+0.72}
& 3.27 & \textbf{4.90} & \delt{+1.63}
& 25.55 & 24.56 & \delt{-0.99} \\

\quad Qwen\(3.5\) (\(27\)B) 
& 8.41 & 8.41 & \delt{0.00}
& 10.63 & \textbf{11.17} & \delt{+0.54}
& 46.70 & \textbf{47.69} & \delt{+0.99} \\

\bottomrule
\end{tabular}%
}
\end{table}
\begin{table}
\centering
\caption{Evaluation on GPQA, MMLU-Pro, and BIG-Bench Hard. $\Delta$: Post minus Direct, in accuracy points.}
\label{tab:phase1_ood}
\resizebox{0.78\textwidth}{!}{%
\begin{tabular}{lccccccccc}
\toprule
& \multicolumn{3}{c}{\textbf{GPQA}} & \multicolumn{3}{c}{\textbf{MMLU-Pro}} & \multicolumn{3}{c}{\textbf{BIG-Bench Hard}} \\
\cmidrule(lr){2-4} \cmidrule(lr){5-7} \cmidrule(lr){8-10}
\textbf{Model} & Direct & Post & $\Delta$ & Direct & Post & $\Delta$ & Direct & Post & $\Delta$ \\
\midrule

\multicolumn{10}{l}{\textbf{Llama Family}} \\
\quad Llama-\(3.1\) (\(8\)B) 
& 29.66 & \textbf{32.81} & \delt{+3.15}
& 27.47 & \textbf{37.17} & \delt{+9.70}
& 42.84 & \textbf{45.69} & \delt{+2.85} \\

\quad Llama-\(3.3\) (\(70\)B) 
& 49.89 & 49.21 & \delt{-0.68}
& 52.33 & \textbf{53.37} & \delt{+1.04}
& 65.32 & \textbf{66.52} & \delt{+1.20} \\

\midrule
\multicolumn{10}{l}{\textbf{Gemma Family}} \\
\quad Gemma-\(3\) (\(12\)B) 
& 33.71 & \textbf{34.61} & \delt{+0.90}
& 44.03 & \textbf{44.17} & \delt{+0.14}
& 59.10 & 58.88 & \delt{-0.22} \\

\quad Gemma-\(3\) (\(27\)B) 
& 36.18 & 34.16 & \delt{-2.02}
& 51.20 & \textbf{51.33} & \delt{+0.13}
& 62.74 & \textbf{63.83} & \delt{+1.09} \\

\midrule
\multicolumn{10}{l}{\textbf{Mistral \& Ministral}} \\
\quad Ministral-\(3\) (\(8\)B) 
& 36.18 & \textbf{37.98} & \delt{+1.80}
& 48.20 & 48.03 & \delt{-0.17}
& 54.86 & \textbf{55.21} & \delt{+0.35} \\

\quad Ministral-\(3\) (\(14\)B) 
& 37.08 & \textbf{39.78} & \delt{+2.70}
& 54.23 & \textbf{54.57} & \delt{+0.34}
& 56.98 & \textbf{58.88} & \delt{+1.90} \\

\quad Mistral-Small (\(24\)B) 
& 37.75 & \textbf{40.22} & \delt{+2.47}
& 53.33 & \textbf{54.43} & \delt{+1.10}
& 59.88 & \textbf{60.80} & \delt{+0.92} \\

\midrule
\multicolumn{10}{l}{\textbf{Qwen Family}} \\
\quad Qwen\(3.5\) (\(4\)B) 
& 37.53 & \textbf{42.92} & \delt{+5.39}
& 43.87 & \textbf{45.57} & \delt{+1.70}
& 52.00 & \textbf{54.42} & \delt{+2.42} \\

\quad Qwen\(3.5\) (\(9\)B) 
& 43.60 & \textbf{58.65} & \delt{+15.05}
& 50.60 & \textbf{51.17} & \delt{+0.57}
& 56.07 & \textbf{57.52} & \delt{+1.45} \\

\quad Qwen\(3.5\) (\(27\)B) 
& 49.89 & \textbf{62.70} & \delt{+12.81}
& 66.00 & 65.87 & \delt{-0.13}
& 69.91 & 68.67 & \delt{-1.24} \\

\bottomrule
\end{tabular}%
}
\end{table}

\subsection{Post Reasoning Effect on Proprietary Models}

We further evaluate post-reasoning on proprietary frontier models, including Gemini-2.5-Flash, Claude-Haiku-4.5, and GPT-5.4-Mini, across the same set of benchmarks (\Cref{tab:api_amc,tab:api_math_combined,tab:api_ood}). Post-reasoning consistently improves performance across a majority of settings, demonstrating that its benefits are not limited to open-source models and generalize to commercially deployed systems.

On competition mathematics benchmarks (AMC and HMMT), post-reasoning yields consistent gains across all models, with improvements observed across nearly all tasks (e.g., up to \textbf{+4.43 points} on AMC 12 for Claude-Haiku-4.5 and \textbf{+3.61 points} on HMMT Feb). These results reinforce earlier observations that post-reasoning is particularly effective for long-horizon, multi-step reasoning tasks.

On GSM8K, gains are smaller and less consistent, mirroring trends observed with open models, with improvements limited to \textbf{+0.99 points} (Claude-Haiku-4.5) and \textbf{+0.37 points} (GPT-5.4-Mini), and a slight decrease for Gemini-2.5-Flash. 

On broader reasoning benchmarks (GPQA, MMLU-Pro, BIG-Bench Hard), post-reasoning improves performance in most cases, with moderate gains (e.g., \textbf{+7.87 points} on GPQA for Gemini-2.5-Flash and \textbf{+9.73 points} on BIG-Bench Hard for Claude-Haiku-4.5). Similar to earlier results, improvements are more pronounced on reasoning-intensive tasks (GPQA) compared to more knowledge-oriented benchmarks.

These results demonstrate that \textbf{post-reasoning generalizes across both open and proprietary models}, consistently improving performance, with the largest gains observed on tasks requiring deeper reasoning.

\begin{table}[ht]
\centering
\caption{Proprietary models on AMC benchmarks. $\Delta$: Post minus Direct, in accuracy points.}
\label{tab:api_amc}
\resizebox{0.78\textwidth}{!}{%
\begin{tabular}{lccccccccc}
\toprule
& \multicolumn{3}{c}{\textbf{AMC 8}} & \multicolumn{3}{c}{\textbf{AMC 10}} & \multicolumn{3}{c}{\textbf{AMC 12}} \\
\cmidrule(lr){2-4} \cmidrule(lr){5-7} \cmidrule(lr){8-10}
\textbf{Model} & Direct & Post & $\Delta$ & Direct & Post & $\Delta$ & Direct & Post & $\Delta$ \\
\midrule

Gemini-2.5-Flash 
& 44.03 & \textbf{47.39} & \delt{+3.36}
& 29.66 & \textbf{32.36} & \delt{+2.70}
& 22.88 & \textbf{24.35} & \delt{+1.47} \\

Claude-Haiku-4.5 
& 35.07 & \textbf{37.31} & \delt{+2.24}
& 27.19 & \textbf{28.99} & \delt{+1.80}
& 19.56 & \textbf{23.99} & \delt{+4.43} \\

GPT-5.4-Mini 
& 31.72 & \textbf{32.84} & \delt{+1.12}
& 21.80 & \textbf{22.92} & \delt{+1.12}
& 15.50 & \textbf{16.61} & \delt{+1.11} \\

\bottomrule
\end{tabular}%
}
\end{table}

\begin{table}[ht]
\centering
\caption{Proprietary models on HMMT and GSM8K. $\Delta$: Post minus Direct, in accuracy points.}
\label{tab:api_math_combined}
\resizebox{0.78\textwidth}{!}{%
\begin{tabular}{lccccccccc}
\toprule
& \multicolumn{3}{c}{\textbf{HMMT Feb}} & \multicolumn{3}{c}{\textbf{HMMT Nov}} & \multicolumn{3}{c}{\textbf{GSM8K}} \\
\cmidrule(lr){2-4} \cmidrule(lr){5-7} \cmidrule(lr){8-10}
\textbf{Model} & Direct & Post & $\Delta$ & Direct & Post & $\Delta$ & Direct & Post & $\Delta$ \\
\midrule

Gemini-2.5-Flash 
& 10.82 & \textbf{11.30} & \delt{+0.48}
& 10.63 & \textbf{11.44} & \delt{+0.81}
& 60.58 & 60.27 & \delt{-0.31} \\

Claude-Haiku-4.5 
& 13.22 & \textbf{16.83} & \delt{+3.61}
& 15.80 & \textbf{16.62} & \delt{+0.82}
& 48.90 & \textbf{49.89} & \delt{+0.99} \\

GPT-5.4-Mini 
& 6.49 & 6.49 & \delt{0.00}
& 7.90 & \textbf{8.17} & \delt{+0.27}
& 52.62 & \textbf{52.99} & \delt{+0.37} \\

\bottomrule
\end{tabular}%
}
\end{table}

\begin{table}[ht]
\centering
\caption{Proprietary models on GPQA, MMLU-Pro, and BIG-Bench Hard. $\Delta$: Post minus Direct, in accuracy points.}
\label{tab:api_ood}
\resizebox{0.78\textwidth}{!}{%
\begin{tabular}{lccccccccc}
\toprule
& \multicolumn{3}{c}{\textbf{GPQA}} & \multicolumn{3}{c}{\textbf{MMLU-Pro}} & \multicolumn{3}{c}{\textbf{BIG-Bench Hard}} \\
\cmidrule(lr){2-4} \cmidrule(lr){5-7} \cmidrule(lr){8-10}
\textbf{Model} & Direct & Post & $\Delta$ & Direct & Post & $\Delta$ & Direct & Post & $\Delta$ \\
\midrule

Gemini-2.5-Flash 
& 49.21 & \textbf{57.08} & \delt{+7.87}
& 72.07 & \textbf{76.13} & \delt{+4.06}
& 73.46 & \textbf{73.77} & \delt{+0.31} \\

Claude-Haiku-4.5 
& 38.20 & \textbf{43.60} & \delt{+5.40}
& 54.73 & \textbf{59.90} & \delt{+5.17}
& 60.87 & \textbf{70.60} & \delt{+9.73} \\

GPT-5.4-Mini 
& 43.82 & \textbf{44.72} & \delt{+0.90}
& 54.47 & 53.30 & \delt{-1.17}
& 60.56 & \textbf{62.88} & \delt{+2.32} \\

\bottomrule
\end{tabular}%
}
\end{table}

\section{Supervised Post-Reason Tuning}

Similar to inherently (pre-)reasoning models~\citep{yang2025qwen3technicalreport,openai2025gptoss120bgptoss20bmodel,deepseekai2025deepseekr1incentivizingreasoningcapability}, where augmented reasoning behavior is internalized as part of the model, post-reasoning can also be learned as an intrinsic capability. To train the model, we use the supervised loss function described in \Cref{sec:loss} and explore three training strategies: (1) \textbf{Expert Distillation}, where the base model is trained on post-reasoning traces generated by a stronger expert model; (2) \textbf{Rephrased Distillation}, where the base model rewrites expert-generated reasoning traces in its own vocabulary; and (3) \textbf{Self-Distillation}, where the base model generates its own explanation conditioned on the input and final answer.

\subsection{Post-Reasoning Train Samples}
\begin{wraptable}{r}{0.48\textwidth}
\centering
\caption{Post-Reason train set, covering a range of reasoning depths from standard algebra to advanced multi-step logic.}
\label{tab:dataset_composition}
\resizebox{\linewidth}{!}{%
\begin{tabular}{llcc}
\toprule
\textbf{Data Source} & \textbf{Domain} & \textbf{Count} & \textbf{\%} \\
\midrule
Orca Math & Algebraic Logic & 1,234 & 35.26 \\
Synthetic Math & General Math & 1,121 & 32.03 \\
CN K-12 & Intermediate & 668 & 19.09 \\
Olympiads & Multi-Step Logic & 447 & 12.77 \\
Synthetic AMC & Competition & 30 & 0.86 \\
\midrule
\textbf{Total} & & \textbf{3,500} & \textbf{100.00} \\
\bottomrule
\end{tabular}%
}
\vspace{2pt}
\caption{Downstream accuracy ablation (\% points) on Mistral-Small (24B). Self-distillation significantly outperforms expert and rephrased traces, particularly on complex reasoning tasks (GPQA).}
\label{tab:ablation_accuracy}
\resizebox{\linewidth}{!}{%
\begin{tabular}{lcccc}
\toprule
\textbf{Training Distribution} & \textbf{MATH} & \textbf{GPQA} & \textbf{MMLU-Pro} & \textbf{BBH} \\
\midrule
Expert Distillation & 26.50 & 36.40 & 54.00 & 59.53 \\
Rephrased Distillation & 26.18 & 38.43 & 54.03 & 59.50 \\
\textbf{Self-Distillation} & \textbf{27.21} & \textbf{45.17} & \textbf{54.53} & \textbf{59.94} \\
\bottomrule
\end{tabular}%
}
\end{wraptable}
To prevent test set contamination, we construct a strictly filtered training corpus of approximately \(3{,}500\) integer-answer mathematical problems from the Numina dataset \cite{numina_math_datasets}, removing any overlap with evaluation benchmarks such as GSM8K, AMC, and HMMT. To promote generalization rather than narrow overfitting, the dataset is curated to span multiple cognitive strata, ranging from standard algebraic reasoning (e.g., Orca Math \cite{mitra2024orcamathunlockingpotentialslms}) to advanced multi-step logic (see Table~\ref{tab:dataset_composition}).
To ensure high-quality reasoning supervision, we prompt models to generate formal, proof-style justifications without prematurely revealing the final answer. Generated traces are filtered to remove instances that leak target assumptions or exhibit insufficient depth (e.g., fewer than 20 words), with invalid samples discarded and regenerated.
\subsection{Training Setup}
All models are fine-tuned using LoRA \citep{hu2021loralowrankadaptationlarge, xu2023parameterefficientfinetuningmethodspretrained} with $r=16$, $\alpha=32$, and dropout $0.05$. Training uses an effective batch size of $32$ for 3 epochs, a learning rate of $2 \times 10^{-5}$, and a cosine scheduler \cite{loshchilov2017sgdrstochasticgradientdescent}. The masked post-reasoning objective (Eq.~\ref{eq:loss_function}) exhibits stable convergence (Figure~\ref{fig:overall-numina-loss}).

As shown in Table~\ref{tab:ablation_accuracy}, self-distillation consistently outperforms both expert and rephrased distillation, with the largest gains observed on GPQA. A likely reason is distributional alignment: in self-distillation, the model generates justifications in its own linguistic and reasoning style, making the resulting reasoning traces more compatible with its internal representations. In contrast, expert distillation introduces a distribution shift, as the model is trained to imitate reasoning traces produced by a different (often larger) model, which may not align with its own inductive biases. Rephrased distillation partially mitigates this mismatch, but still depends on external traces.

The advantage of self-distillation is particularly pronounced on complex reasoning tasks such as GPQA, where generating coherent, multi-step justifications is critical. Here, internally consistent reasoning appears more beneficial than imitating externally generated traces. On broader benchmarks (MMLU-Pro, BBH), the gains are smaller, suggesting that when tasks rely more on knowledge recall than deep reasoning, the choice of reasoning distribution has a reduced impact. Given its consistently strong performance and better alignment with the model’s native reasoning distribution, we adopt self-distillation for all subsequent post-reason training experiments.

\begin{table*}[t]
\centering
\caption{Post-Reason training results on mathematical reasoning benchmarks. Each cell reports Post-Reason (PR) / Post-Reason SFT accuracy, with SFT minus PR, in accuracy points, in subscript.}
\label{tab:phase2_math_compact}
\resizebox{\textwidth}{!}{%
\begin{tabular}{lcccccc}
\toprule
\textbf{Model} 
& \textbf{AMC 8} 
& \textbf{AMC 10} 
& \textbf{AMC 12} 
& \textbf{HMMT Feb} 
& \textbf{HMMT Nov} 
& \textbf{GSM8K} \\
\midrule

\multicolumn{7}{l}{\textbf{Llama Family}} \\
Llama-3.1 (8B) 
& 11.57 / \textbf{14.18}$_{(+2.61)}$
& 12.13 / \textbf{12.58}$_{(+0.45)}$
& 10.70 / \textbf{12.55}$_{(+1.85)}$
& 2.88 / \textbf{3.61}$_{(+0.73)}$
& 2.18 / \textbf{2.72}$_{(+0.54)}$
& 13.65 / 13.12$_{(-0.53)}$ \\

Llama-3.3 (70B) 
& 28.73 / 28.73$_{(0.00)}$
& 33.03 / \textbf{33.26}$_{(+0.23)}$
& 26.94 / \textbf{27.68}$_{(+0.74)}$
& 9.86 / \textbf{11.78}$_{(+1.92)}$
& 10.35 / 9.81$_{(-0.54)}$
& 35.78 / \textbf{36.01}$_{(+0.23)}$ \\

\midrule
\multicolumn{7}{l}{\textbf{Gemma Family}} \\
Gemma-3 (12B) 
& 17.16 / \textbf{24.63}$_{(+7.47)}$
& 13.26 / \textbf{13.71}$_{(+0.45)}$
& 6.64 / \textbf{9.23}$_{(+2.59)}$
& 4.33 / \textbf{4.81}$_{(+0.48)}$
& 5.99 / 5.72$_{(-0.27)}$
& 21.68 / 20.55$_{(-1.13)}$ \\

Gemma-3 (27B) 
& 15.30 / \textbf{18.28}$_{(+2.98)}$
& 17.08 / 16.63$_{(-0.45)}$
& 15.13 / 12.55$_{(-2.58)}$
& 6.25 / \textbf{6.97}$_{(+0.72)}$
& 7.90 / \textbf{8.17}$_{(+0.27)}$
& 31.77 / \textbf{31.92}$_{(+0.15)}$ \\

\midrule
\multicolumn{7}{l}{\textbf{Mistral \& Ministral}} \\
Ministral-3 (8B) 
& 18.66 / 17.91$_{(-0.75)}$
& 11.69 / \textbf{12.81}$_{(+1.12)}$
& 7.75 / \textbf{11.44}$_{(+3.69)}$
& 4.57 / 3.37$_{(-1.20)}$
& 4.36 / \textbf{4.63}$_{(+0.27)}$
& 21.15 / \textbf{23.96}$_{(+2.81)}$ \\

Ministral-3 (14B) 
& 21.27 / \textbf{23.88}$_{(+2.61)}$
& 15.51 / \textbf{16.18}$_{(+0.67)}$
& 9.59 / \textbf{9.96}$_{(+0.37)}$
& 4.33 / \textbf{5.05}$_{(+0.72)}$
& 5.18 / \textbf{6.27}$_{(+1.09)}$
& 27.22 / \textbf{27.29}$_{(+0.07)}$ \\

Mistral-Small (24B) 
& 13.81 / \textbf{16.79}$_{(+2.98)}$
& 13.48 / \textbf{16.63}$_{(+3.15)}$
& 7.75 / \textbf{9.23}$_{(+1.48)}$
& 3.61 / \textbf{4.57}$_{(+0.96)}$
& 7.08 / \textbf{7.90}$_{(+0.82)}$
& 29.95 / \textbf{30.63}$_{(+0.68)}$ \\

\midrule
\multicolumn{7}{l}{\textbf{Qwen Family}} \\
Qwen3.5 (4B) 
& 20.90 / \textbf{22.39}$_{(+1.49)}$
& 10.79 / \textbf{12.36}$_{(+1.57)}$
& 5.17 / \textbf{6.27}$_{(+1.10)}$
& 3.37 / \textbf{5.29}$_{(+1.92)}$
& 3.54 / 3.54$_{(0.00)}$
& 19.48 / \textbf{20.24}$_{(+0.76)}$ \\

Qwen3.5 (9B) 
& 21.12 / 20.94$_{(-0.18)}$
& 16.63 / \textbf{16.85}$_{(+0.22)}$
& 9.59 / \textbf{11.44}$_{(+1.85)}$
& 4.33 / \textbf{4.57}$_{(+0.24)}$
& 4.90 / \textbf{6.81}$_{(+1.91)}$
& 24.56 / \textbf{26.08}$_{(+1.52)}$ \\

Qwen3.5 (27B) 
& 30.67 / \textbf{32.35}$_{(+1.68)}$
& 32.36 / \textbf{34.16}$_{(+1.80)}$
& 20.66 / \textbf{22.51}$_{(+1.85)}$
& 8.41 / \textbf{8.89}$_{(+0.48)}$
& 11.17 / \textbf{11.44}$_{(+0.27)}$
& 47.69 / 47.46$_{(-0.23)}$ \\

\bottomrule
\end{tabular}%
}
\end{table*}

\subsection{Results}

\begin{wraptable}{r}{0.55\linewidth}
\vspace{-10pt}
\centering
 \caption{Post-Reason training results on cross-domain reasoning benchmarks. Format as in Table~\ref{tab:phase2_math_compact}.}
\label{tab:phase2_general_compact}
\scriptsize
\resizebox{0.55\textwidth}{!}{%
\begin{tabular}{lccc}
\toprule
\textbf{Model} & \textbf{GPQA} & \textbf{MMLU-Pro} & \textbf{BBH} \\
\midrule

\multicolumn{4}{l}{\textbf{Llama}} \\
L3.1 (8B) & 32.81 / \textbf{34.16}$_{(+1.35)}$ & 37.17 / 36.80$_{(-0.37)}$ & 45.69 / 43.68$_{(-2.01)}$ \\
L3.3 (70B) & 49.21 / \textbf{50.11}$_{(+0.90)}$ & 53.37 / 53.13$_{(-0.24)}$ & 66.52 / 65.97$_{(-0.55)}$ \\

\multicolumn{4}{l}{\textbf{Gemma}} \\
G3 (12B) & 34.61 / 34.16$_{(-0.45)}$ & 44.17 / \textbf{44.57}$_{(+0.40)}$ & 58.88 / \textbf{59.27}$_{(+0.39)}$ \\
G3 (27B) & 34.16 / \textbf{36.40}$_{(+2.24)}$ & 51.33 / 51.27$_{(-0.06)}$ & 63.83 / \textbf{65.47}$_{(+1.64)}$ \\

\multicolumn{4}{l}{\textbf{Mistral}} \\
M3 (8B) & 37.98 / 35.51$_{(-2.47)}$ & 48.03 / 47.97$_{(-0.06)}$ & 55.21 / \textbf{55.64}$_{(+0.43)}$ \\
M3 (14B) & 39.78 / \textbf{40.22}$_{(+0.44)}$ & 54.57 / 53.60$_{(-0.97)}$ & 58.88 / \textbf{60.42}$_{(+1.54)}$ \\
MS (24B) & 40.22 / \textbf{45.17}$_{(+4.95)}$ & 54.43 / \textbf{54.53}$_{(+0.10)}$ & 60.80 / 59.94$_{(-0.86)}$ \\

\multicolumn{4}{l}{\textbf{Qwen}} \\
Q3.5 (4B) & 42.92 / \textbf{43.37}$_{(+0.45)}$ & 45.57 / \textbf{45.67}$_{(+0.10)}$ & 54.42 / 53.99$_{(-0.43)}$ \\
Q3.5 (9B) & 58.65 / \textbf{60.90}$_{(+2.25)}$ & 51.17 / \textbf{51.50}$_{(+0.33)}$ & 57.52 / \textbf{58.47}$_{(+0.95)}$ \\
Q3.5 (27B) & 62.70 / \textbf{63.60}$_{(+0.90)}$ & 65.87 / 65.53$_{(-0.34)}$ & 68.67 / \textbf{68.78}$_{(+0.11)}$ \\

\bottomrule
\end{tabular}
}
\vspace{-10pt}
\end{wraptable}

\paragraph{Mathematical Reasoning.}

As shown in Table~\ref{tab:phase2_math_compact}, supervised post-reason tuning further improves performance over prompt-based post-reasoning across a large majority of model--benchmark combinations. The gains are particularly pronounced on competition mathematics benchmarks such as AMC and HMMT, where multi-step and long-horizon reasoning are critical. In several cases, gains exceed \textbf{2 points} (up to \textbf{+7.47}), especially for smaller and mid-sized models, indicating that post-reasoning behavior can be effectively internalized through supervised training.

The largest improvements are consistently observed on AMC and HMMT, suggesting that post-reason tuning strengthens structured reasoning capabilities beyond what is achieved through prompting alone. In contrast, gains on GSM8K are comparatively smaller and occasionally negative, likely because GSM8K relies on shorter reasoning chains where prompt-based post-reasoning is already sufficient, leaving limited room for additional improvement through training.

\paragraph{Cross-domain Generalization.}
Table~\ref{tab:phase2_general_compact} shows that the benefits of post-reason tuning extend beyond mathematical tasks. Despite being trained exclusively on mathematical reasoning data, the tuned models outperform direct answering in 25 of 30 settings and prompt-based post-reasoning in 18 of 30 settings across GPQA, MMLU-Pro, and BIG-Bench Hard, demonstrating effective cross-domain transfer.

Improvements are most notable on GPQA, where gains are positive for 8 of 10 models and occasionally substantial (e.g., up to \textbf{+4.95 points}), reflecting its reliance on multi-step reasoning and uncertainty resolution. In contrast, changes on MMLU-Pro and BBH are smaller, within \textbf{$\pm$1} and \textbf{$\pm$2.1} points respectively, with gains for 4 of 10 and 6 of 10 models. This aligns with the nature of these benchmarks, which rely more on knowledge retrieval and shallow reasoning compared to GPQA.

\section{Post-Reasoning vs. Generic Instruction Augmentation}
\begin{wraptable}{r}{0.48\textwidth}
\vspace{-10pt}
\centering
\caption{Instruction augmentations. Values denote average accuracy (\%).}
\label{tab:ablation_post_task}
\resizebox{0.47\textwidth}{!}{
\begin{tabular}{lccc}
\toprule
\textbf{Benchmark} & \textbf{Post-Summary} & \textbf{Post-Confidence} & \textbf{Post-Reason} \\
\midrule
GPQA & 32.59 & 34.53 & \textbf{35.88} \\
MMLU-Pro & 43.54 & 44.40 & \textbf{45.26} \\
GSM8K & 20.24 & 21.05 & \textbf{21.76} \\
\bottomrule
\end{tabular}
}
\vspace{-10pt}
\end{wraptable}
A natural question is whether the observed gains stem from generic instruction augmentation or if explicit reasoning is uniquely beneficial. As shown in Table~\ref{tab:ablation_post_task}, the cognitive objective of the post-generation task is the primary driver of performance gains.

Specifically, the \textbf{Post-Summary} variant consistently underperforms because summarization merely restates the problem and the selected output. The \textbf{Post-Confidence} variant offers a stronger baseline by encouraging metacognitive reflection, yet it remains inferior to \textbf{Post-Reasoning}. By explicitly requiring the model to justify its output, \textbf{Post-Reasoning} imposes a causal constraint between the answer and subsequent tokens. As models compute information for future tokens at the current position~\citep{wu2024planahead}, a required justification shapes the computation of the answer itself. Masked training follows this path: with loss only on the justification, the answer still improves (Tables~\ref{tab:phase2_math_compact}--\ref{tab:phase2_general_compact}). The gain is thus not a byproduct of token count.

\section{Limitations}
While supervised post-reason tuning improves direct-answer capabilities, it has several limitations. First, performance on tasks requiring deep algorithmic search, such as HMMT, remains bounded, as complex logic trees may still benefit from explicit Chain-of-Thought (CoT) reasoning \cite{wei2022chain, snell2024scalingllmtesttimecompute}. Second, gains on widely represented benchmarks such as GSM8K are comparatively small due to already strong direct-inference capabilities from extensive pretraining exposure. In some cases, additional reasoning can even introduce minor regressions, consistent with recent findings on reasoning redundancy in simple mathematical tasks \cite{chen2024not, srivastava2026llms, aggarwal2025optimalthinkingbench}. Finally, although self-distillation avoids human-annotated reasoning traces by relying only on question-answer pairs, its effectiveness depends on task complexity, with shallow factual tasks providing weaker supervision signals than multi-step reasoning problems.

\section{Conclusion}
We introduced Post-Reasoning, which conditions models to justify answers after generating them, with no reasoning tokens before the answer. Across 13 models, it improves direct-answer accuracy in 84.62\% of settings, and supervised post-reason tuning embeds this behavior into the weights, improving 91.11\% of settings.

\clearpage

{\raggedright
\bibliographystyle{plainnat}
\bibliography{references}
}
\clearpage

\appendix
\section*{Appendix Contents}

\newcommand{\appcontentsline}[2]{%
    \noindent
    \hyperref[#1]{Appendix~\ref*{#1}: #2}%
    \dotfill
    \hyperref[#1]{\pageref*{#1}}%
    \par
}

\appcontentsline{app:when_to_think}
{When to Use Thinking \& How to Decide?}

\appcontentsline{app:inference_details}
{Detailed Experimental Setup}

\appcontentsline{app:sft_implementation}
{Supervised Fine-Tuning (SFT) Setup}

\appcontentsline{app:thinking_results}
{Native Latent Thinking Prompting Evaluations}

\appcontentsline{app:latency_results}
{Answer Latency and Parsing Reliability}

\appcontentsline{sec:appendix_ablation}
{Post-Task Ablation}

\appcontentsline{app:extended_optimization}
{Extended Optimization Dynamics}

\appcontentsline{app:dataset_ablation}
{Training-Corpus Dependency}

\clearpage

\section{When to Use Thinking \& How to Decide?}
\label{app:when_to_think}

\subsection{How to switch between thinking and non-thinking?}
In practice, one of the most promising approaches is to employ a \textbf{router model} that predicts whether a query requires reasoning and routes it to a non-thinking or thinking model accordingly (e.g., as shown in Table 3 of \citealp{aggarwal2025optimalthinkingbench}). A router can be obtained through training-free methods \citep{su2026cprouter,tang2026vdarrouter} or trained to predict whether a query will benefit from additional reasoning. For instance, the approach in \citet{guo2026thinkwhenneeded} trains such a router by comparing the performance and token cost of Think and Non-Think modes and learning the expected utility gain of reasoning. During inference, reasoning is invoked only when the predicted gain exceeds a predefined threshold or budget.

\subsection{Which tasks?}
OptimalThinkingBench \citep{aggarwal2025optimalthinkingbench} provides a broad characterization of which tasks benefit from substantial reasoning versus minimal reasoning. Additional reasoning is generally beneficial for tasks involving algorithms, graph reasoning, arithmetic, geometry, formal logic, and strategic games. In contrast, simple factual retrieval, general knowledge questions, and other straightforward queries often derive little or no benefit from thinking and may even experience degraded performance due to unnecessary reasoning. OptimalThinkingBench comprises 72 domains in which thinking provides little or no performance benefit. Determining the appropriate amount of thinking for a given task has therefore emerged as an increasingly important research direction \citep{fang2025thinkless,jiang2025thinkonly,lou2025adacot,zhang2025continue,yu2025thinksmarter,liang2025thinkswitcher,luo2025adar1,zhang2025adaptthink,chen2024not}.

Overall, although thinking models are essential for reasoning-intensive tasks, \textbf{non-thinking models remain equally important}, making model routing a practical solution for dynamically selecting between thinking and non-thinking models.

\clearpage

\section{Detailed Experimental Setup}
\label{app:inference_details}
To ensure full transparency and reproducibility, this section provides comprehensive details regarding the hardware, software, decoding configurations, models, and prompt templates utilized in our prompting and supervised post-reason tuning experiments, including both localized open-source deployments and proprietary API integrations.

\subsection{Hardware Infrastructure}
All baseline inference evaluations for open-source weights were conducted on a localized high-performance compute cluster equipped with NVIDIA H200 Tensor Core GPUs, each featuring 141GB of VRAM. Due to the high memory capacity of the H200 accelerators, inference for models up to 32B parameters was efficiently executed on single-GPU nodes. The only exception was \texttt{Llama-3.3-70B-Instruct}, which required tensor parallelism distributed across 2x H200 GPUs to accommodate the massive model weights and extended KV-cache. Conversely, evaluations involving proprietary closed-source models were conducted via their respective official APIs, bypassing these local hardware dependencies.

\subsection{Software Stack and Libraries}
The experimental pipeline was built upon a robust Python software stack, structurally organized to follow the chronological progression of our methodology:

\begin{itemize}
    \item \textbf{Prompting (Inference \& Evaluation):} All localized open-source model generation was accelerated using the \texttt{vllm} engine \citep{kwon2023efficient}. Proprietary model evaluations were orchestrated asynchronously utilizing the \texttt{openai} Python SDK and \texttt{asyncio}, with concurrency tracked via \texttt{tqdm}. Downstream evaluation, regular expression answer parsing, and result aggregation were executed using \texttt{pandas}, \texttt{json}, and \texttt{re}.
    \item \textbf{Supervised Post-Reason Tuning (Distillation Data Generation):} The synthetic corpora required for Target-Conditioned Self-Distillation were generated, filtered, and concatenated utilizing the \texttt{openai} Python SDK, \texttt{asyncio}, and the Hugging Face \texttt{datasets} library \citep{lhoest2021datasetscommunitylibrarynatural}.
    \item \textbf{Supervised Post-Reason Tuning (Model Training):} Model weights, tokenization, and training loops were implemented using PyTorch \citep{paszke2019pytorchimperativestylehighperformance} and the Hugging Face \texttt{transformers} library \citep{wolf2020huggingfacestransformersstateoftheartnatural}. Parameter-efficient fine-tuning via Low-Rank Adaptation (LoRA) was configured using \texttt{peft}.
\end{itemize}

\subsection{Inference Engine and API Configuration}
To maintain consistent high-throughput generation across the diverse open-source model suite, we deployed the vLLM engine \cite{kwon2023efficient}. A dedicated \texttt{tmux} orchestration pipeline was used to spin up independent vLLM servers mapped to specific network ports for asynchronous client requests.

To optimize the KV-cache and prevent Out-of-Memory (OOM) errors during varying contextual loads, the \texttt{--max-model-len} argument was dynamically scaled:
\begin{itemize}
    \item \textbf{Standard Instruct Evaluation:} Capped at \(16,384\) tokens.
    \item \textbf{Native Thinking Evaluation:} Expanded to \(32,768\) tokens to accommodate the extensive, unbounded latent reasoning traces generated by models like GPT-OSS, Qwen3 and Qwen3.5.
\end{itemize}

For Qwen-family models leveraging latent planning, the \texttt{--reasoning-parser qwen3} flag was explicitly passed. All local models utilized \texttt{--enable-prefix-caching} to accelerate the 3-shot prompt processing, and GPU memory utilization was strictly bounded at $0.90$. 

For the closed-source proprietary models (Gemini-2.5-Flash, Claude-Haiku-4.5, and GPT-5.4-Mini), request orchestration was managed via customized asynchronous Python pipelines interfacing with their official REST APIs. To ensure strict parity with the standard instruct baselines and isolate the effect of structural prompting, all native latent reasoning features (e.g., Anthropic's \texttt{thinking} blocks, OpenAI's \texttt{reasoning\_effort}, and Gemini's \texttt{thinking\_budget}) were explicitly disabled during generation.

\subsection{Model Deployment Registry}
Our evaluation pipeline utilized a diverse suite of models distributed across the experimental stages. In the \textbf{Prompting Evaluation (Inference-Time Prompting Efficacy)}, all 14 distinct open-source models were evaluated to establish rigorous zero-shot baselines. This included the standard instruct models as well as the GPT-OSS 20B, Qwen3, and Qwen3.5 families, which were additionally evaluated on their thinking capabilities. Furthermore, a \textbf{Prompting Extension} incorporated proprietary models to provide state-of-the-art closed-source baselines.

In the \textbf{Supervised Post-Reason Tuning Evaluation}, the focus shifted to evaluating the architecture weights updated via the SFT pipeline. This stage was conducted specifically on the standard instruct architectures (Llama, Gemma, Ministral, Mistral) and the Qwen3.5 family. Notably, the Qwen3 and GPT-OSS models were utilized exclusively during prompt-based evaluations for native thinking comparisons and were not subjected to the supervised post-reason tuning pipeline. Table \ref{tab:model_registry} details the complete registry alongside their specific deployment protocols.

\begin{table}[htbp]
    \centering
    \caption{Complete Model Deployment Registry (Prompting \& SFT)}
    \label{tab:model_registry}
    \resizebox{0.96\textwidth}{!}{%
    \begin{tabular}{l l c c c}
        \toprule
        \textbf{Model Family} & \textbf{Model Identifier} & \textbf{Experiment} & \textbf{GPU Config} & \textbf{Port} \\
        \midrule
        \multirow{2}{*}{Gemma 3} 
        & \texttt{gemma-3-12b-it} & Prompting, SFT & 1x H200 & 8010 \\
        & \texttt{gemma-3-27b-it} & Prompting, SFT & 1x H200 & 8011 \\
        \midrule
        \multirow{2}{*}{Llama 3} 
        & \texttt{Llama-3.1-8B-Instruct} & Prompting, SFT & 1x H200 & 8018 \\
        & \texttt{Llama-3.3-70B-Instruct} & Prompting, SFT & 2x H200 & 8009 \\
        \midrule
        \multirow{2}{*}{Ministral} 
        & \texttt{Ministral-3-8B-Instruct-2512} & Prompting, SFT & 1x H200 & 8015 \\
        & \texttt{Ministral-3-14B-Instruct-2512} & Prompting, SFT & 1x H200 & 8016 \\
        \midrule
        Mistral & \texttt{Mistral-Small-24B-Instruct-2501} & Prompting, SFT & 1x H200 & 8017 \\
        \midrule
        \multirow{3}{*}{Qwen 3} 
        & \texttt{Qwen3-8B} & Prompting & 1x H200 & 8012 \\
        & \texttt{Qwen3-14B} & Prompting & 1x H200 & 8013 \\
        & \texttt{Qwen3-32B} & Prompting & 1x H200 & 8014 \\
        \midrule
        \multirow{3}{*}{Qwen 3.5} 
        & \texttt{Qwen35-4B} & Prompting, SFT & 1x H200 & 8019 \\
        & \texttt{Qwen35-9B} & Prompting, SFT & 1x H200 & 8020 \\
        & \texttt{Qwen35-27B} & Prompting, SFT & 1x H200 & 8021 \\
        \midrule
        GPT-OSS & \texttt{GPT-OSS-20B} & Prompting & 1x H200 & 8008 \\
        \midrule
        \multicolumn{5}{l}{\textbf{Proprietary API Models (Prompting Extension)}} \\
        \midrule
        Gemini & \texttt{gemini-2.5-flash} & Prompting & API & - \\
        Claude & \texttt{claude-haiku-4-5} & Prompting & API & - \\
        GPT & \texttt{gpt-5.4-mini} & Prompting & API & - \\
        \bottomrule
    \end{tabular}%
    }
\end{table}

\subsection{Generation Hyperparameters}
Hyperparameters were standardized based on the architectural capabilities of the models and the requirements of the active prompting strategy. For standard instruct baselines in both prompting and supervised post-reason tuning evaluations, generation was constrained to \(2,048\) tokens for direct extraction and \(4,096\) tokens for post-reason extraction on local models. All evaluations used temperature 0.0. For the proprietary API models, maximum output tokens were uniformly expanded to \(8,192\) to ensure responses were not prematurely truncated during extensive post-reasoning generations. 

For the auxiliary thinking evaluations during prompt-based testing, the evaluated open-source models (GPT-OSS, Qwen3, Qwen3.5) utilized native reasoning tags. Consequently, their token limits were expanded significantly to \(16,384\) tokens. Table \ref{tab:hyperparams_detailed} provides the exact decoding parameters used across the different prompting strategies.

\begin{table}[htbp]
    \centering
    \caption{Exact Generation Hyperparameters by Architecture and Strategy}
    \label{tab:hyperparams_detailed}
    \resizebox{\textwidth}{!}{%
    \begin{tabular}{l l c c c c c c}
        \toprule
        \textbf{Model Class} & \textbf{Strategy} & \textbf{Max Tokens} & \textbf{Temp.} & \textbf{Top-p} & \textbf{Top-k} & \textbf{Pres. Pen.} & \textbf{Rep. Pen.} \\
        \midrule
        \multirow{2}{*}{\shortstack[l]{Llama, Ministral, Mistral, Gemma\\(Prompting \& SFT)}} 
        & Direct Answer & 2,048 & 0.0 & 0.8 & 20 & - & - \\
        & Post-Reason & 4,096 & 0.0 & 0.8 & 20 & - & - \\
        \midrule
        \multirow{4}{*}{\shortstack[l]{GPT-OSS, Qwen 3\\(Prompting Only)}} 
        & Direct Answer & 2,048 & 0.0 & 0.8 & 20 & - & - \\
        & Post-Reason & 4,096 & 0.0 & 0.8 & 20 & - & - \\
        & Thinking Direct & 16,384 & 0.0 & 0.95 & 20 & - & - \\
        & Thinking Post & 16,384 & 0.0 & 0.95 & 20 & - & - \\
        \midrule
        \multirow{4}{*}{\shortstack[l]{Qwen 3.5\\(Prompting \& SFT)}} 
        & Direct Answer & 2,048 & 0.0 & 0.8 & 20 & 1.5 & 1.0 \\
        & Post-Reason & 4,096 & 0.0 & 0.8 & 20 & 1.5 & 1.0 \\
        & Thinking Direct & 16,384 & 0.0 & 0.95 & 20 & 1.5 & 1.0 \\
        & Thinking Post & 16,384 & 0.0 & 0.95 & 20 & 1.5 & 1.0 \\
        \midrule
        \multirow{4}{*}{\shortstack[l]{Proprietary APIs\\(Prompting Extension)}} 
        & Direct & 2,048 & 0.0 & - & - & - & - \\
        & Post-Reason & 4,096 & 0.0 & - & - & - & - \\
        & Thinking Direct & 16,384 & 0.0 & - & - & - & - \\
        & Thinking Post & 16,384 & 0.0 & - & - & - & - \\
        \bottomrule
    \end{tabular}%
    }
\end{table}

\subsection{Prompting Frameworks}
\label{app:prompts_all}
All evaluations used \textbf{3-shot prompting}. For each benchmark, the placeholder \texttt{\textless task-specific instruction\textgreater} comprises the benchmark system instruction, the 3-shot examples, and the question. We then append the Direct or Post-Reason suffix to this complete task-specific instruction. The Thinking Direct suffix is identical to the Direct suffix, and Thinking Post is identical to the Post-Reason suffix. The two prompts therefore have the following form:

\begin{center}
\fbox{%
\begin{minipage}{0.94\linewidth}
\textbf{Direct/ Thinking Direct Prompt}\\[3pt]
\ttfamily\raggedright
\textless task-specific instruction\textgreater\ State the final answer immediately.
\end{minipage}%
}

\vspace{6pt}

\fbox{%
\begin{minipage}{0.94\linewidth}
\textbf{Post-Reason/ Thinking Post Prompt}\\[3pt]
\ttfamily\raggedright
\textless task-specific instruction\textgreater\ State the final answer immediately and justify your answer.
\end{minipage}%
}
\end{center}

As outlined in our methodology, these strategies were deployed in prompt-based evaluation to establish distinct comparative baselines:
\begin{itemize}
    \item \textbf{Standard Prompting (All Models):} Evaluated using the \textbf{Direct Answer} and \textbf{Post-Reason} system instructions, completely disabling internal thinking to test standard generative formatting.
    \item \textbf{Native Thinking (Select Models):} Restricted to the GPT-OSS 20B, Qwen3, and Qwen3.5 model families for open-source models. For closed-source models, it is restricted to Gemini-2.5-Flash, Claude-Haiku-4.5, and GPT-5.4-Mini, with thinking level set to low. These were evaluated using the \textbf{Thinking Direct} and \textbf{Thinking Post} instructions, which contain identical textual constraints to the standard prompts but with thinking to invoke the internal \texttt{<think>} reasoning block prior to generating visible text.
\end{itemize}

Tables \ref{tab:benchmark_system_prompts} detail the exact system prompts  utilized for each respective benchmark in our suite.

\begin{table}[htbp]
    \centering
    \caption{Generic benchmark system instructions shared by Direct, Post-Reason, Thinking Direct, and Thinking Post.}
    \label{tab:benchmark_system_prompts}
    \begin{tabular}{p{0.19\linewidth} p{0.74\linewidth}}
        \toprule
        \textbf{Benchmark} & \textbf{System instruction} \\
        \midrule
        AMC 8 & Solve the competition mathematics problem. Return the final answer as an integer. \\
        AMC 10 & Solve the competition mathematics problem. Return the final answer as an integer. \\
        AMC 12 & Solve the competition mathematics problem. Return the final answer as an integer. \\
        HMMT February & Solve the competition mathematics problem. Return the final answer as an integer. \\
        HMMT November & Solve the competition mathematics problem. Return the final answer as an integer. \\
        GSM8K & Solve the grade-school mathematics problem. Return the final answer as a number. \\
        GPQA & Answer the graduate-level science multiple-choice question. Return the letter of the correct option. \\
        MMLU-Pro & Answer the multiple-choice question. Return one option letter from A through J. \\
        BBH & Solve the task and return the answer in the format required by the question. \\
        \bottomrule
    \end{tabular}
\end{table}

\clearpage

\section{Supervised Fine-Tuning (SFT) Setup}
\label{app:sft_implementation}

To explicitly embed the post-reasoning cognitive pathway into the model weights, we utilized a highly constrained Supervised Fine-Tuning (SFT) pipeline. Rather than training the model on the full conversational sequence, we implemented a \textbf{Targeted Loss Masking} algorithm to prevent the model from overfitting to the instruction prompts or memorizing the answers. 

During the tokenization phase, the system prompt, the user query, and the immediate statement of the final numerical answer were explicitly masked by assigning them a label index of \texttt{-100}. Consequently, the autoregressive cross-entropy loss was computed \textit{exclusively} over the generative tokens corresponding to the detailed logical trajectory (the \texttt{Explanation:} block). This forced the parameter updates to optimize solely for the structural and logical integrity of the latent reasoning, entirely decoupling the reasoning process from the prompt structure.

\subsection{Computational Cost and Resource Allocation}
To provide full transparency regarding the computational footprint and environmental impact of our study, we highly optimized the Supervised Fine-Tuning (SFT) pipeline to minimize both hardware requirements and human oversight. 

Due to the memory efficiency of LoRA combined with DeepSpeed ZeRO-3 and Flash Attention 2, the complete fine-tuning process was successfully executed using only \textbf{1x NVIDIA H200 (141GB) GPU per model}. The sole exception to this single-node constraint was the \texttt{Llama-3.3-70B} architecture, which required distribution across \textbf{2x NVIDIA H200 GPUs} to safely accommodate the expanded optimizer states and gradient checkpoints without triggering Out-of-Memory (OOM) failures. Furthermore, the streamlined data processing and training orchestration required approximately \textbf{2 hours of dedicated setup and monitoring per model}, demonstrating the practical viability of our targeted masking approach for rapid model alignment.

\subsection{Hyperparameter Configuration}
All fine-tuning was executed using Low-Rank Adaptation (LoRA) \cite{hu2021loralowrankadaptationlarge} to efficiently update the attention and multi-layer perceptron (MLP) blocks. To maintain high memory efficiency across our H200 cluster, we utilized gradient checkpointing, Flash Attention 2 \cite{dao2023flashattention2fasterattentionbetter}, and bfloat16 precision. The exact training and LoRA parameters utilized across all models are detailed in Table \ref{tab:sft_hyperparams_exact}.

\begin{table}[htbp]
    \centering
    \caption{Exact SFT and LoRA Hyperparameters}
    \label{tab:sft_hyperparams_exact}
    \begin{tabular}{l l}
        \toprule
        \textbf{Hyperparameter} & \textbf{Value} \\
        \midrule
        \multicolumn{2}{c}{LoRA Configuration} \\
        \midrule
        Rank ($r$) & 16 \\
        Alpha ($\alpha$) & 32 \\
        Dropout & 0.05 \\
        Target Modules & \shortstack[l]{\texttt{q\_proj, k\_proj, v\_proj, o\_proj,} \\ \texttt{gate\_proj, up\_proj, down\_proj}} \\
        \midrule
        \multicolumn{2}{c}{Training Arguments} \\
        \midrule
        Learning Rate & $2 \times 10^{-5}$ \\
        LR Scheduler & Cosine \\
        Warmup Ratio & 0.1 \\
        Epochs & 3 \\
        Per-Device Batch Size & 2 \\
        Gradient Accumulation Steps & 16 \\
        Max Sequence Length & 4,096 \\
        Optimizer & AdamW (\texttt{adamw\_torch}) \\
        Precision & bfloat16 \\
        Gradient Checkpointing & Enabled \\
        \bottomrule
    \end{tabular}
\end{table}

\subsection{Answer-Token Masking Ablation}
\label{app:masking_ablation}

We compared the masked Post-Reason objective against an unmasked objective that applies loss to both the answer and justification. The data, LoRA configuration, and seeds were identical; the loss mask over the answer was the only difference. Table~\ref{tab:masking_ablation} reports the existing three-seed results. Masking the answer produced higher observed accuracy in all nine evaluated settings, showing that direct supervision of the answer tokens was not required for these gains.

\begin{table}[htbp]
    \centering
    \caption{Post-Reason SFT accuracy under masked and unmasked objectives. Values are mean accuracy over three seeds. $\Delta$ is Masked minus Unmasked in accuracy points.}
    \label{tab:masking_ablation}
    \resizebox{\textwidth}{!}{%
    \begin{tabular}{lccccccccc}
        \toprule
        & \multicolumn{3}{c}{\textbf{AMC 10}} & \multicolumn{3}{c}{\textbf{GPQA}} & \multicolumn{3}{c}{\textbf{MMLU-Pro}} \\
        \cmidrule(lr){2-4} \cmidrule(lr){5-7} \cmidrule(lr){8-10}
        \textbf{Model} & Unmasked & Masked & $\Delta$ & Unmasked & Masked & $\Delta$ & Unmasked & Masked & $\Delta$ \\
        \midrule
        \multicolumn{10}{l}{\textbf{Qwen-3.5 Family}} \\
        \quad Qwen-3.5 (4B) & 9.75 & 10.86 & $+1.11^{*}$ & 40.90 & 42.87 & $+1.97^{*}$ & 44.90 & 45.70 & $+0.80^{*}$ \\
        \midrule
        \multicolumn{10}{l}{\textbf{Ministral-3 Family}} \\
        \quad Ministral-3 (14B) & 14.51 & 16.01 & $+1.50^{*}$ & 35.51 & 39.68 & $+4.17^{*}$ & 51.43 & 54.91 & $+3.48^{*}$ \\
        \midrule
        \multicolumn{10}{l}{\textbf{Llama-3.3 Family}} \\
        \quad Llama-3.3 (70B) & 32.01 & 33.15 & $+1.14^{*}$ & 48.09 & 49.36 & $+1.27^{*}$ & 51.67 & 53.31 & $+1.64^{*}$ \\
        \bottomrule
    \end{tabular}%
    }
    \vspace{2pt}

    \begin{minipage}{\textwidth}
    \footnotesize $^{*}p<0.05$, one-sided Welch t-test, $n=3$ seeds per arm.
    \end{minipage}
\end{table}

\clearpage

\section{Native Latent Thinking Prompting Evaluations}
\label{app:thinking_results}

While the core experiments established the regularizing effects of post-reasoning constraints on standard generative models, modern architectures are increasingly equipped with native latent reasoning mechanisms (e.g., \texttt{<think>} tokens). To provide a comprehensive analysis of inference-time prompting, we conducted an auxiliary evaluation to determine how structural formatting interacts with these built-in cognitive blocks.

\subsection{Experimental Design and Scope}
This evaluation was strictly bounded to model families in our registry natively trained to support latent reasoning traces: the GPT-OSS (20B), Qwen3 (8B, 14B, 32B), and Qwen3.5 (4B, 9B, 27B) architectures. For closed-source models, the evaluation additionally included Gemini-2.5-Flash, Claude-Haiku-4.5, and GPT-5.4-Mini with thinking level set to low. 

We established two comparative baselines analogous to our standard instruct evaluations, but modified to explicitly invoke the reasoning blocks prior to generating visible text:
\begin{itemize}
    \item \textbf{Thinking Direct:} The model utilizes its \texttt{<think>} block to explore the problem space, followed immediately by generating the final extracted answer.
    \item \textbf{Thinking Post:} The model utilizes its \texttt{<think>} block, outputs the final extracted answer, and is then forced to generate a visible, step-by-step post-reasoning justification.
\end{itemize}

Our objective was to observe whether the structural constraint of generating a post-reasoning explanation provides any supplementary regularization when the model has already been afforded a dedicated, unconstrained latent thinking phase.

\subsection{Thinking Baseline Results and Analysis}
\begin{figure}[h]
    \centering
    \includegraphics[width=0.6\textwidth]{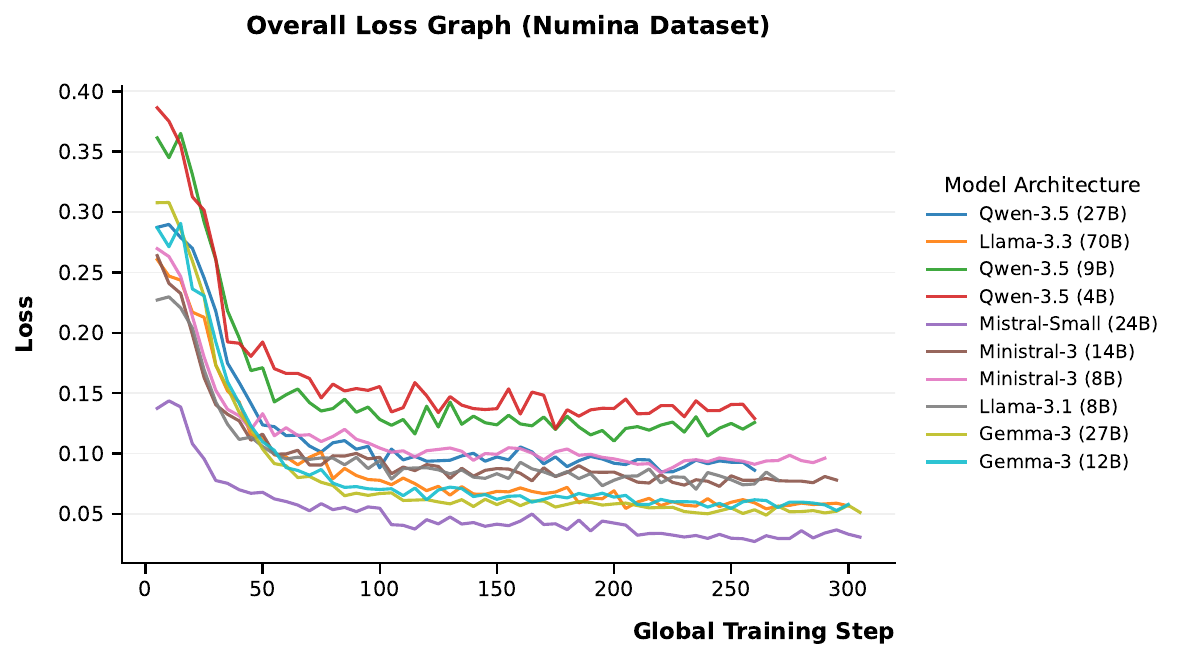} 
    \caption{Training loss curve for the Post-Reason SFT framework, demonstrating stable gradient descent over the masked rationale.}
    \label{fig:overall-numina-loss}
\end{figure}

The addition of native thinking mechanisms allocates more compute and explore logical trajectories prior to outputting the initial answer token, the relative impact of formatting the subsequent text is shifted. Figure \ref{fig:overall-numina-loss} illustrates the overall training loss curve for the Post-Reason SFT framework. 

Tables \ref{tab:think_amc} through \ref{tab:closed_reasoning} detail the performance of the \textbf{Thinking Direct} (denoted in tables as Think Dir) and \textbf{Thinking Post} (denoted in tables as Think Post) strategies across our benchmark suite.

These results reveal a nuanced interplay between latent reasoning and output formatting. For Qwen-3 (8B, 32B), the Think Dir instruction to answer without reasoning leads the model to leave its \texttt{<think>} block empty on part of the competition items (78\% on AMC 8 for Qwen-3 8B), so the gains in these cells partly reflect this instruction. Conversely, highly capable models (e.g., Qwen-3.5 27B) exhibit diminishing returns or slight regressions under formatting, indicating that their latent planning is already optimal and forcing additional output structure can occasionally disrupt their inherent reasoning pathways.

\begin{table}[htbp]
\centering
\caption{Open-source models: AMC Competition Mathematics. $\Delta$ is Think Post minus Think Dir, in accuracy points.}
\label{tab:think_amc}
\resizebox{\textwidth}{!}{%
\begin{tabular}{lccccccccc}
\toprule
& \multicolumn{3}{c}{\textbf{AMC 8}} & \multicolumn{3}{c}{\textbf{AMC 10}} & \multicolumn{3}{c}{\textbf{AMC 12}} \\
\cmidrule(lr){2-4} \cmidrule(lr){5-7} \cmidrule(lr){8-10}
\textbf{Model} & Think Dir & Think Post & $\Delta$ & Think Dir & Think Post & $\Delta$ & Think Dir & Think Post & $\Delta$ \\
\midrule
\multicolumn{10}{l}{\textbf{GPT-OSS Family}} \\
\quad GPT-OSS (20B) & 86.19 & 90.30 & +4.11 & 84.27 & 87.42 & +3.15 & 79.34 & 83.03 & +3.69 \\
\midrule
\multicolumn{10}{l}{\textbf{Qwen-3 Family}} \\
\quad Qwen-3 (8B) & 23.51 & 91.79 & +68.28 & 75.96 & 88.54 & +12.58 & 60.52 & 76.01 & +15.49 \\
\quad Qwen-3 (14B) & 92.54 & 94.78 & +2.24 & 91.01 & 90.56 & -0.45 & 83.03 & 79.70 & -3.33 \\
\quad Qwen-3 (32B) & 85.82 & 86.19 & +0.37 & 87.19 & 81.57 & -5.62 & 72.69 & 71.22 & -1.47 \\
\midrule
\multicolumn{10}{l}{\textbf{Qwen-3.5 Family}} \\
\quad Qwen-3.5 (4B) & 80.97 & 94.03 & +13.06 & 74.83 & 89.89 & +15.06 & 50.92 & 86.35 & +35.43 \\
\quad Qwen-3.5 (9B) & 74.63 & 93.28 & +18.65 & 79.10 & 90.11 & +11.01 & 46.49 & 77.49 & +31.00 \\
\quad Qwen-3.5 (27B) & 92.16 & 90.67 & -1.49 & 93.26 & 88.54 & -4.72 & 82.29 & 78.23 & -4.06 \\
\bottomrule
\end{tabular}%
}
\end{table}

\begin{table}[htbp]
\centering
\caption{Open-source models: Standard and Competition Mathematics. $\Delta$ is Think Post minus Think Dir, in accuracy points.}
\label{tab:think_combined_math}
\resizebox{\textwidth}{!}{%
\begin{tabular}{l ccc ccc ccc}
\toprule
& \multicolumn{3}{c}{\textbf{GSM8K}} & \multicolumn{3}{c}{\textbf{HMMT Feb}} & \multicolumn{3}{c}{\textbf{HMMT Nov}} \\
\cmidrule(lr){2-4} \cmidrule(lr){5-7} \cmidrule(lr){8-10}
\textbf{Model} & Think Dir & Think Post & $\Delta$ & Think Dir & Think Post & $\Delta$ & Think Dir & Think Post & $\Delta$ \\
\midrule
\multicolumn{10}{l}{\textbf{GPT-OSS Family}} \\
\quad GPT-OSS (20B) & 95.53 & 95.83 & +0.30 & 50.72 & 54.33 & +3.61 & 62.67 & 64.31 & +1.64 \\
\midrule
\multicolumn{10}{l}{\textbf{Qwen-3 Family}} \\
\quad Qwen-3 (8B) & 95.00 & 95.83 & +0.83 & 28.85 & 43.99 & +15.14 & 27.52 & 59.13 & +31.61 \\
\quad Qwen-3 (14B) & 95.75 & 96.36 & +0.61 & 50.72 & 50.24 & -0.48 & 64.31 & 64.31 & 0.00 \\
\quad Qwen-3 (32B) & 94.47 & 97.04 & +2.57 & 18.51 & 46.15 & +27.64 & 32.97 & 59.13 & +26.16 \\
\midrule
\multicolumn{10}{l}{\textbf{Qwen-3.5 Family}} \\
\quad Qwen-3.5 (4B) & 95.83 & 96.36 & +0.53 & 11.06 & 48.08 & +37.02 & 31.06 & 59.67 & +28.61 \\
\quad Qwen-3.5 (9B) & 97.04 & 97.12 & +0.08 & 12.50 & 38.70 & +26.20 & 29.97 & 57.22 & +27.25 \\
\quad Qwen-3.5 (27B) & 97.42 & 97.42 & 0.00 & 35.34 & 25.72 & -9.62 & 54.22 & 41.96 & -12.26 \\
\bottomrule
\end{tabular}%
}
\end{table}

\begin{table}[htbp]
\centering
\caption{Open-source models: General Reasoning Benchmarks. $\Delta$ is Think Post minus Think Dir, in accuracy points.}
\label{tab:think_ood}
\resizebox{\textwidth}{!}{%
\begin{tabular}{lccccccccc}
\toprule
& \multicolumn{3}{c}{\textbf{GPQA}} & \multicolumn{3}{c}{\textbf{MMLU-Pro}} & \multicolumn{3}{c}{\textbf{BIG-Bench Hard}} \\
\cmidrule(lr){2-4} \cmidrule(lr){5-7} \cmidrule(lr){8-10}
\textbf{Model} & Think Dir & Think Post & $\Delta$ & Think Dir & Think Post & $\Delta$ & Think Dir & Think Post & $\Delta$ \\
\midrule
\multicolumn{10}{l}{\textbf{GPT-OSS Family}} \\
\quad GPT-OSS (20B) & 60.00 & 60.90 & +0.90 & 72.90 & 74.47 & +1.57 & 88.60 & 92.20 & +3.60 \\
\midrule
\multicolumn{10}{l}{\textbf{Qwen-3 Family}} \\
\quad Qwen-3 (8B) & 51.91 & 53.71 & +1.80 & 75.33 & 75.33 & 0.00 & 89.42 & 89.91 & +0.49 \\
\quad Qwen-3 (14B) & 52.58 & 55.96 & +3.38 & 77.97 & 78.20 & +0.23 & 89.91 & 91.49 & +1.58 \\
\quad Qwen-3 (32B) & 57.98 & 59.55 & +1.57 & 80.17 & 80.83 & +0.66 & 91.61 & 92.27 & +0.66 \\
\midrule
\multicolumn{10}{l}{\textbf{Qwen-3.5 Family}} \\
\quad Qwen-3.5 (4B) & 66.07 & 68.54 & +2.47 & 78.50 & 79.23 & +0.73 & 90.68 & 92.84 & +2.16 \\
\quad Qwen-3.5 (9B) & 72.81 & 73.93 & +1.12 & 82.10 & 82.27 & +0.17 & 91.78 & 93.81 & +2.03 \\
\quad Qwen-3.5 (27B) & 77.98 & 81.12 & +3.14 & 87.07 & 86.50 & -0.57 & 95.61 & 95.53 & -0.08 \\
\bottomrule
\end{tabular}%
}
\end{table}

\clearpage 

\subsection{Closed-Source Models}
\label{app:low_effort_results}

We evaluated Gemini-2.5-Flash, Claude-Haiku-4.5, and GPT-5.4-Mini under four conditions. Direct and Post-Reason used native thinking disabled. Thinking Direct used exactly the Direct prompt with thinking level set to low, while Thinking Post used exactly the Post-Reason prompt with thinking level set to low. All four conditions used the same system instruction, three-shot prefix, question, model setting, and temperature-zero decoding.

Tables~\ref{tab:closed_amc}--\ref{tab:closed_reasoning} report accuracy and differences in accuracy points. Post-Reason improved all 27 low-effort model--benchmark settings, with a mean improvement of +4.26 points and a median improvement of +3.22 points. The observed improvements range from +1.35 to +15.86 points.

\begin{table}[htbp]
\centering
\caption{Closed-source models: AMC Competition Mathematics. $\Delta$ is the second arm minus the first (Post-Reason minus Direct, or Think Post minus Think Dir), in accuracy points.}
\label{tab:closed_amc}
\resizebox{\textwidth}{!}{%
\begin{tabular}{lccccccccc}
\toprule
& \multicolumn{3}{c}{\textbf{AMC 8}} & \multicolumn{3}{c}{\textbf{AMC 10}} & \multicolumn{3}{c}{\textbf{AMC 12}} \\
\cmidrule(lr){2-4} \cmidrule(lr){5-7} \cmidrule(lr){8-10}
\textbf{Model} & Direct & Post-Reason & $\Delta$ & Direct & Post-Reason & $\Delta$ & Direct & Post-Reason & $\Delta$ \\
\midrule
Claude-Haiku-4.5 & 35.07 & 37.31 & +2.24 & 27.19 & 28.99 & +1.80 & 19.56 & 23.99 & +4.43 \\
GPT-5.4-Mini & 31.72 & 32.84 & +1.12 & 21.80 & 22.92 & +1.12 & 15.50 & 16.61 & +1.11 \\
Gemini-2.5-Flash & 44.03 & 47.39 & +3.36 & 29.66 & 32.36 & +2.70 & 22.88 & 24.35 & +1.47 \\
\midrule
& \multicolumn{3}{c}{\textbf{AMC 8}} & \multicolumn{3}{c}{\textbf{AMC 10}} & \multicolumn{3}{c}{\textbf{AMC 12}} \\
\cmidrule(lr){2-4} \cmidrule(lr){5-7} \cmidrule(lr){8-10}
\textbf{Model} & Think Dir & Think Post & $\Delta$ & Think Dir & Think Post & $\Delta$ & Think Dir & Think Post & $\Delta$ \\
\midrule
Claude-Haiku-4.5 & 82.84 & 89.18 & +6.34 & 72.36 & 79.33 & +6.97 & 57.93 & 67.16 & +9.23 \\
GPT-5.4-Mini & 87.03 & 89.55 & +2.52 & 80.40 & 84.49 & +4.09 & 79.34 & 81.55 & +2.21 \\
Gemini-2.5-Flash & 76.87 & 78.36 & +1.49 & 59.27 & 61.57 & +2.30 & 43.76 & 46.13 & +2.37 \\
\bottomrule
\end{tabular}%
}
\end{table}

\begin{table}[htbp]
\centering
\caption{Closed-source models: Standard and Competition Mathematics. $\Delta$ is the second arm minus the first (Post-Reason minus Direct, or Think Post minus Think Dir), in accuracy points.}
\label{tab:closed_math}
\resizebox{\textwidth}{!}{%
\begin{tabular}{lccccccccc}
\toprule
& \multicolumn{3}{c}{\textbf{GSM8K}} & \multicolumn{3}{c}{\textbf{HMMT Feb}} & \multicolumn{3}{c}{\textbf{HMMT Nov}} \\
\cmidrule(lr){2-4} \cmidrule(lr){5-7} \cmidrule(lr){8-10}
\textbf{Model} & Direct & Post-Reason & $\Delta$ & Direct & Post-Reason & $\Delta$ & Direct & Post-Reason & $\Delta$ \\
\midrule
Claude-Haiku-4.5 & 48.90 & 49.89 & +0.99 & 13.22 & 16.83 & +3.61 & 15.80 & 16.62 & +0.82 \\
GPT-5.4-Mini & 52.62 & 52.99 & +0.37 & 6.49 & 6.49 & 0.00 & 7.90 & 8.17 & +0.27 \\
Gemini-2.5-Flash & 60.58 & 60.27 & -0.31 & 10.82 & 11.30 & +0.48 & 10.63 & 11.44 & +0.81 \\
\midrule
& \multicolumn{3}{c}{\textbf{GSM8K}} & \multicolumn{3}{c}{\textbf{HMMT Feb}} & \multicolumn{3}{c}{\textbf{HMMT Nov}} \\
\cmidrule(lr){2-4} \cmidrule(lr){5-7} \cmidrule(lr){8-10}
\textbf{Model} & Think Dir & Think Post & $\Delta$ & Think Dir & Think Post & $\Delta$ & Think Dir & Think Post & $\Delta$ \\
\midrule
Claude-Haiku-4.5 & 95.04 & 97.74 & +2.70 & 28.61 & 44.47 & +15.86 & 50.95 & 53.95 & +3.00 \\
GPT-5.4-Mini & 91.30 & 95.30 & +4.00 & 50.96 & 54.33 & +3.37 & 61.85 & 65.40 & +3.55 \\
Gemini-2.5-Flash & 90.91 & 93.63 & +2.72 & 13.94 & 15.62 & +1.68 & 18.80 & 20.16 & +1.36 \\
\bottomrule
\end{tabular}%
}
\end{table}

\begin{table}[htbp]
\centering
\caption{Closed-source models: General Reasoning Benchmarks. $\Delta$ is the second arm minus the first (Post-Reason minus Direct, or Think Post minus Think Dir), in accuracy points.}
\label{tab:closed_reasoning}
\resizebox{\textwidth}{!}{%
\begin{tabular}{lccccccccc}
\toprule
& \multicolumn{3}{c}{\textbf{GPQA}} & \multicolumn{3}{c}{\textbf{MMLU-Pro}} & \multicolumn{3}{c}{\textbf{BBH}} \\
\cmidrule(lr){2-4} \cmidrule(lr){5-7} \cmidrule(lr){8-10}
\textbf{Model} & Direct & Post-Reason & $\Delta$ & Direct & Post-Reason & $\Delta$ & Direct & Post-Reason & $\Delta$ \\
\midrule
Claude-Haiku-4.5 & 38.20 & 43.60 & +5.40 & 54.73 & 59.90 & +5.17 & 60.87 & 70.60 & +9.73 \\
GPT-5.4-Mini & 43.82 & 44.72 & +0.90 & 54.47 & 53.30 & -1.17 & 60.56 & 62.88 & +2.32 \\
Gemini-2.5-Flash & 49.21 & 57.08 & +7.87 & 72.07 & 76.13 & +4.06 & 73.46 & 73.77 & +0.31 \\
\midrule
& \multicolumn{3}{c}{\textbf{GPQA}} & \multicolumn{3}{c}{\textbf{MMLU-Pro}} & \multicolumn{3}{c}{\textbf{BBH}} \\
\cmidrule(lr){2-4} \cmidrule(lr){5-7} \cmidrule(lr){8-10}
\textbf{Model} & Think Dir & Think Post & $\Delta$ & Think Dir & Think Post & $\Delta$ & Think Dir & Think Post & $\Delta$ \\
\midrule
Claude-Haiku-4.5 & 57.53 & 58.88 & +1.35 & 78.07 & 81.90 & +3.83 & 85.38 & 93.80 & +8.42 \\
GPT-5.4-Mini & 65.09 & 68.31 & +3.22 & 78.13 & 81.87 & +3.74 & 87.69 & 91.06 & +3.37 \\
Gemini-2.5-Flash & 57.30 & 68.31 & +11.01 & 82.33 & 83.70 & +1.37 & 90.37 & 93.26 & +2.89 \\
\bottomrule
\end{tabular}%
}
\end{table}

\clearpage

\section{Answer Latency and Parsing Reliability}
\label{app:latency_results}

We measured wall-clock time from request submission to the final answer token under Direct and deployment-time Post-Reason. For Post-Reason, generation stops when the answer field is complete, before generating the requested justification. Measurements used single requests on an idle H200 node; each entry in Table~\ref{tab:answer_latency} is the range of per-benchmark medians across nine benchmarks. Post-Reason remained within 0.02--0.07 seconds of Direct across the three evaluated models.

\begin{table}[htbp]
\centering
\caption{Wall-clock time to the final answer token. Post denotes answer-stopped Post-Reason.}
\label{tab:answer_latency}
\begin{tabular}{lrrr}
\toprule
\textbf{Strategy} & \textbf{Llama-3.1 (8B)} & \textbf{Llama-3.3 (70B)} & \textbf{Qwen-3.5 (9B)} \\
\midrule
Direct & 0.02--0.05 s & 0.07--0.17 s & 0.04--0.08 s \\
Post & 0.05 s & 0.15--0.24 s & 0.06--0.09 s \\
\bottomrule
\end{tabular}
\end{table}

\begin{table}[htbp]
\centering
\caption{Single-stream decoding throughput and median time to first token (TTFT). Ranges show the minimum and maximum across nine benchmarks.}
\label{tab:serving_throughput}
\begin{tabular}{lrr}
\toprule
\textbf{Model} & \textbf{Decode throughput (tokens/s)} & \textbf{Median TTFT} \\
\midrule
Llama-3.1 (8B) & 176 (156--181) & 26 ms \\
Llama-3.3 (70B) & 45 (42--46) & 77 ms \\
Qwen-3.5 (9B) & 169 (156--171) & 40 ms \\
\bottomrule
\end{tabular}
\end{table}

Across the 90 evaluated model--benchmark cells, the answer preceded the justification and was parsed successfully on at least 99.6\% of examples. Median tokens to answer were 10.5 for Direct and 10.9 for Post-Reason. The larger generation limit for Post-Reason applies only when the post-answer justification is retained.

\clearpage

\section{Prompting - Extended Experimental Details: Post-Task Ablation}
\label{sec:appendix_ablation}

In Table \ref{tab:ablation_post_task}, we explored the efficacy of different post-generation tasks to isolate the effect of logical justification. To isolate whether the content generated after the answer matters, we compare three answer-first continuation tasks: a justification, a summary, and a confidence statement. This ablation was conducted across three representative models from our suite: Llama-3.1 (8B), Gemma-3 (12B), and Mistral-Small (24B). To ensure a fair comparison, the same questions, 3-shot examples, answer representation, and decoding configuration are used throughout. Only the requested post-answer content changes. 

Table~\ref{tab:post_task_suffixes} gives the common, benchmark-independent suffixes.

\begin{table}[htbp]
\centering
\caption{Unified suffixes for the post-task ablation.}
\label{tab:post_task_suffixes}
\begin{tabular}{p{0.20\linewidth}>{\raggedright\arraybackslash}p{0.72\linewidth}}
\toprule
\textbf{Condition} & \textbf{Appended suffix} \\
\midrule
\textbf{Post-Reason} & \texttt{State the final answer immediately and justify your answer.} \\
\addlinespace
\textbf{Post-Summary} & \texttt{State the final answer immediately and then provide a concise summary.} \\
\addlinespace
\textbf{Post-Confidence} & \texttt{State the final answer immediately and then report your confidence with a brief explanation.} \\
\bottomrule
\end{tabular}
\end{table}

The corresponding results are reported in Table~\ref{tab:ablation_post_task} in the main paper. Under a comparable answer-first format and continuation budget, Post-Summary underperforms Post-Confidence, which in turn underperforms Post-Reason. This pattern supports the interpretation that the requested justification contributes more than continuation length alone.

\clearpage

\section{Supervised Post-Reason Tuning - Extended Optimization Dynamics}
\label{app:extended_optimization}

To demonstrate that the convergence benefits of self-distillation are not architectural anomalies, we provide the paired training loss curves (Self-Distillation vs. Standard Rephrased Distillation) across the complete suite of fine-tuned models.

\begin{figure}[p] 
    \centering
    
    \begin{minipage}{0.48\textwidth}
        \centering
        \includegraphics[width=0.85\linewidth]{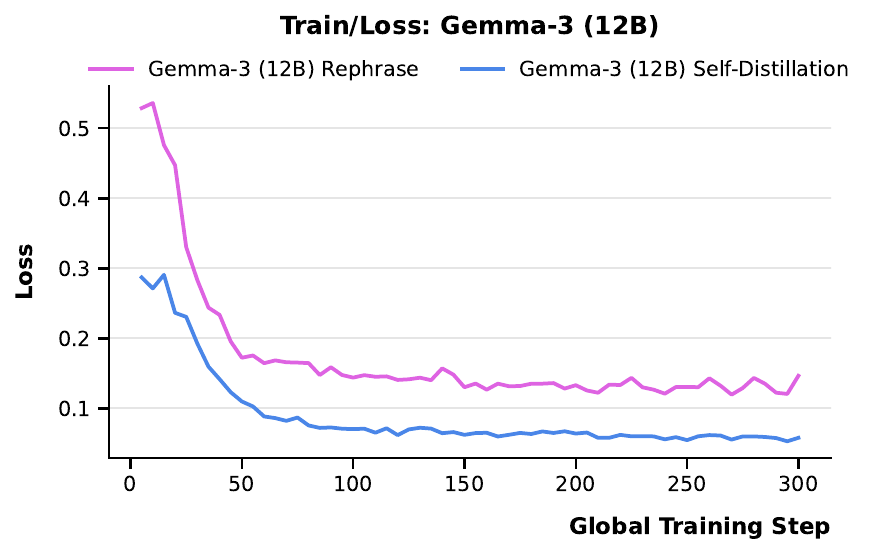}
        
        \vspace{-0.15cm}
        \small (a) Gemma-3 (12B)
    \end{minipage}\hfill
    \begin{minipage}{0.48\textwidth}
        \centering
        \includegraphics[width=0.85\linewidth]{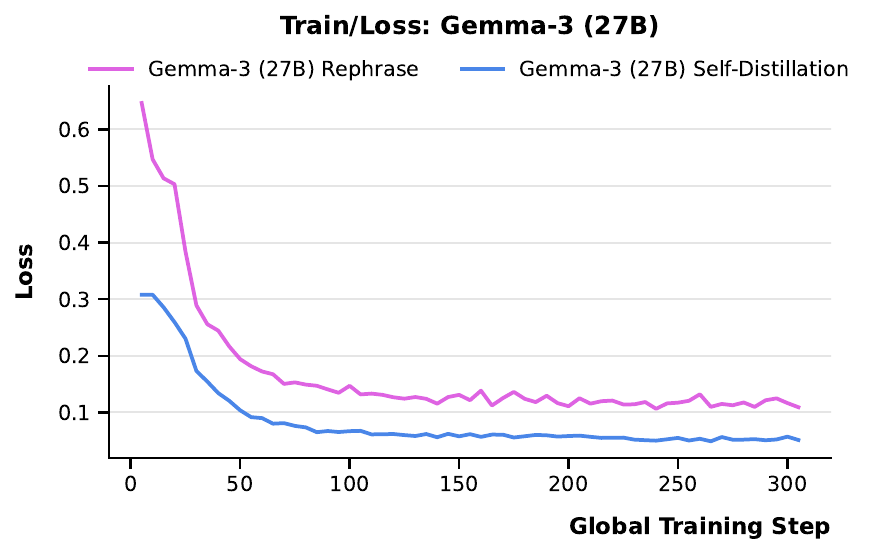}
        
        \vspace{-0.15cm}
        \small (b) Gemma-3 (27B)
    \end{minipage}
    
    \vspace{0.2cm}
    
    \begin{minipage}{0.48\textwidth}
        \centering
        \includegraphics[width=0.85\linewidth]{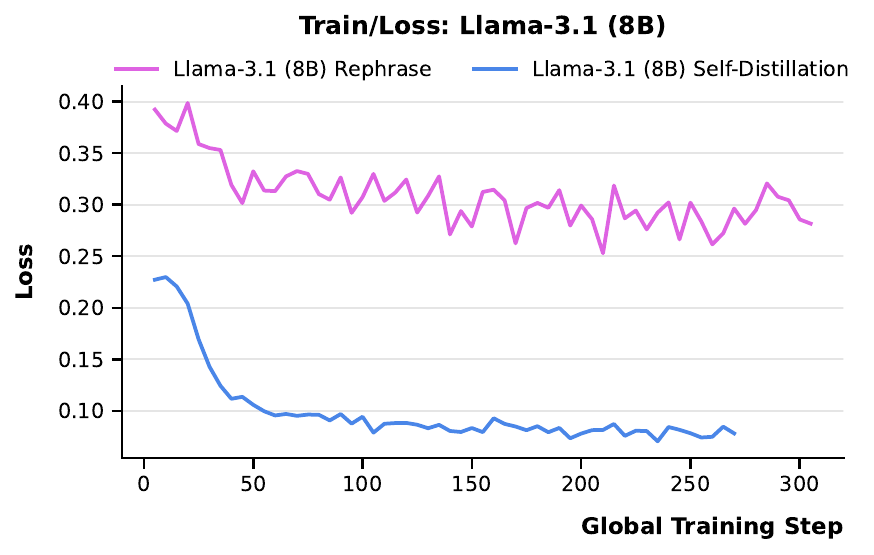}
        
        \vspace{-0.15cm}
        \small (c) Llama-3.1 (8B)
    \end{minipage}\hfill
    \begin{minipage}{0.48\textwidth}
        \centering
        \includegraphics[width=0.85\linewidth]{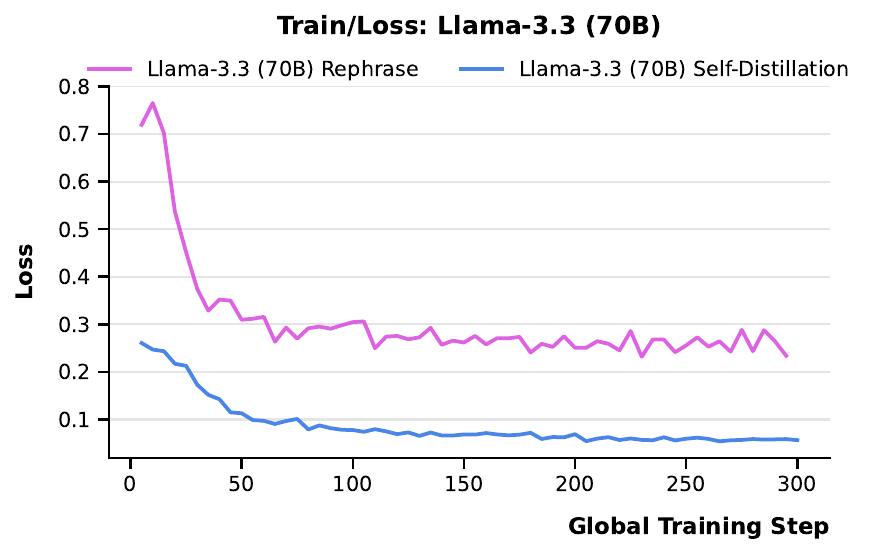}
        
        \vspace{-0.15cm}
        \small (d) Llama-3.3 (70B)
    \end{minipage}
    
    \vspace{0.2cm}
    
    \begin{minipage}{0.48\textwidth}
        \centering
        \includegraphics[width=0.85\linewidth]{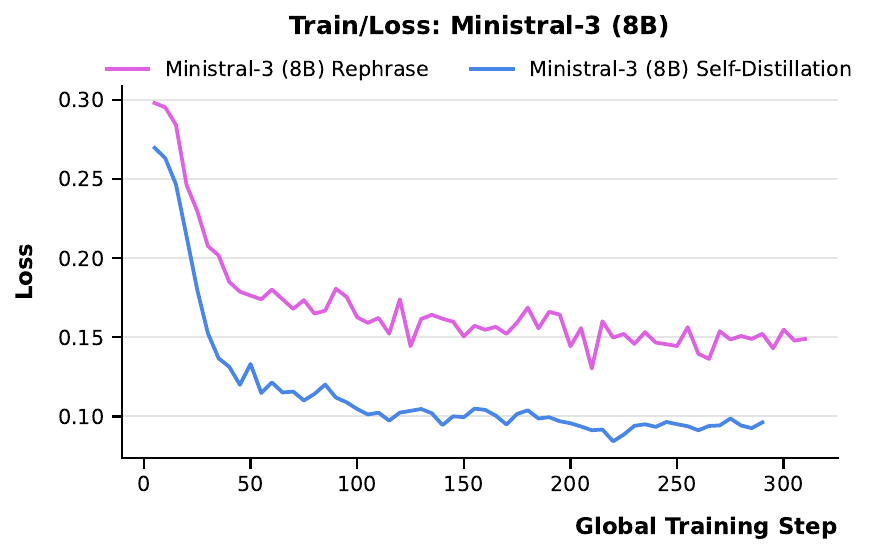}
        
        \vspace{-0.15cm}
        \small (e) Ministral-3 (8B)
    \end{minipage}\hfill
    \begin{minipage}{0.48\textwidth}
        \centering
        \includegraphics[width=0.85\linewidth]{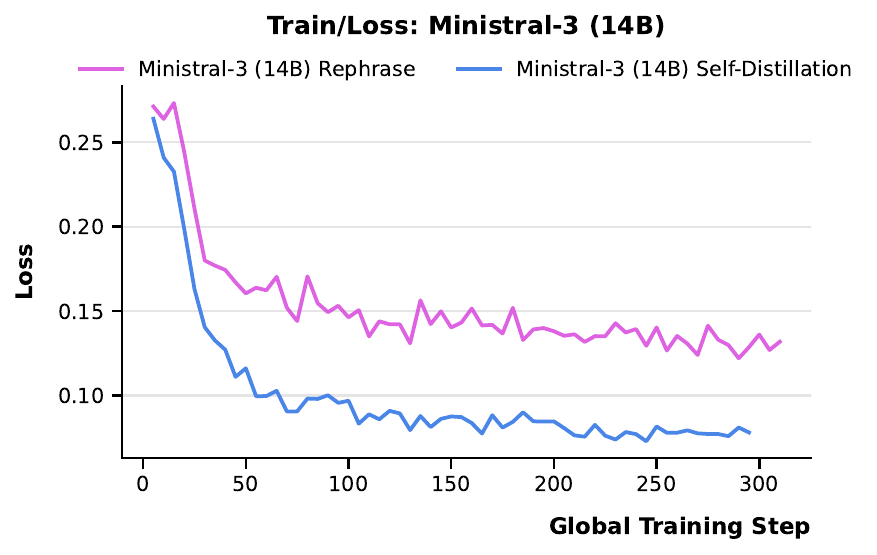}
        
        \vspace{-0.15cm}
        \small (f) Ministral-3 (14B)
    \end{minipage}

    \vspace{0.2cm}
    
    \begin{minipage}{0.48\textwidth}
        \centering
        \includegraphics[width=0.85\linewidth]{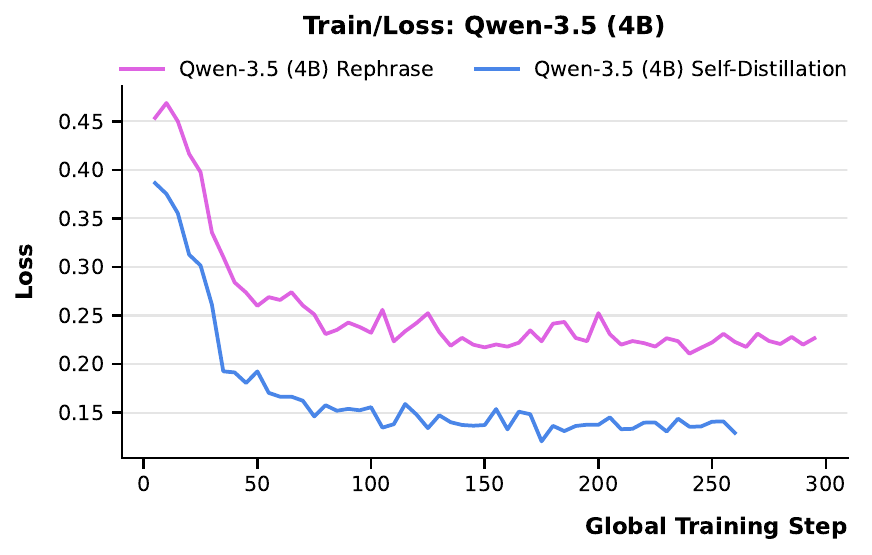}
        
        \vspace{-0.15cm}
        \small (g) Qwen-3.5 (4B)
    \end{minipage}\hfill
    \begin{minipage}{0.48\textwidth}
        \centering
        \includegraphics[width=0.85\linewidth]{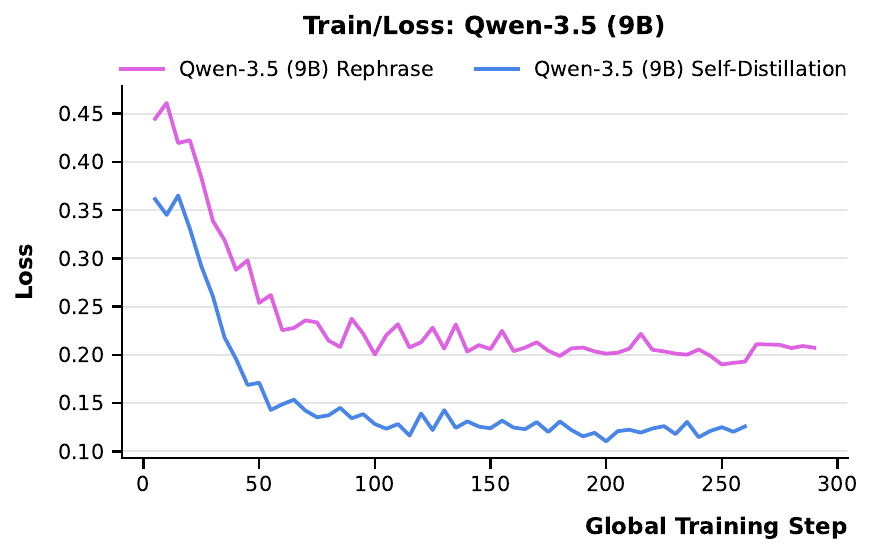}
        
        \vspace{-0.15cm}
        \small (h) Qwen-3.5 (9B)
    \end{minipage}

    \vspace{0.2cm}
    
    \begin{minipage}{0.48\textwidth}
        \centering
        \includegraphics[width=0.85\linewidth]{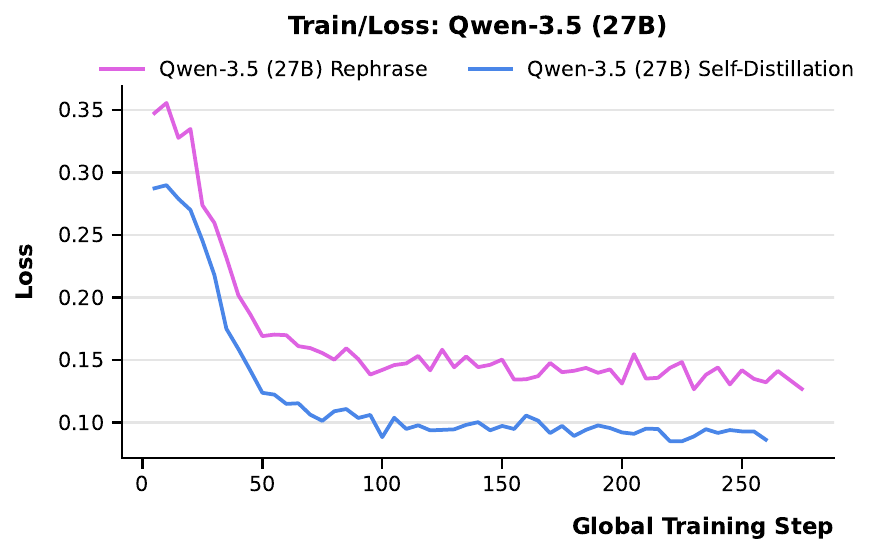}
        
        \vspace{-0.15cm}
        \small (i) Qwen-3.5 (27B)
    \end{minipage}\hfill
    \begin{minipage}{0.48\textwidth}
        \centering
        \includegraphics[width=0.85\linewidth]{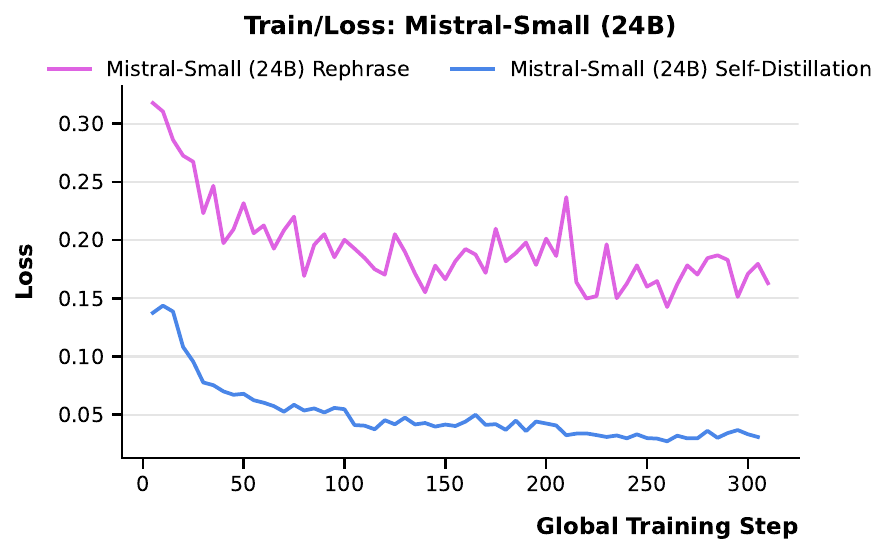}
        
        \vspace{-0.15cm}
        \small (j) Mistral-Small (24B)
    \end{minipage}
    
    \vspace{0.3cm} 
    
    \caption{Extended loss convergence comparisons for all 10 models. Solid blue lines represent the proposed Self-Distillation, while magenta lines represent the baseline Rephrased Distillation. Across all scales and architectures, the proposed method yields lower variance and faster convergence.}
    \label{fig:appendix_loss_grid}
\end{figure}

To ensure a rigorous comparison, all paired training runs were executed utilizing identical hyperparameters (detailed in Appendix \ref{app:sft_implementation}), hardware configurations, and random seeds. As observed in Figure \ref{fig:appendix_loss_grid}, target-conditioned self-distillation yields strictly superior optimization dynamics characterized by three distinct phenomena:

\begin{itemize}
    \item \textbf{Accelerated Initial Convergence:} Across all architectures, the self-distilled objective achieves a lower loss basin significantly faster during the first epoch.
    \item \textbf{Gradient Stability:} The standard rephrased distillation curves exhibit higher variance and periodic spiking, whereas the targeted masking approach produces a distinctly smoother descent trajectory.
    \item \textbf{Scale Invariance:} The smoothing effect and lower-bound convergence hold true regardless of the base model's parameter count (spanning 4B to 70B) or lineage (Gemma, Llama, Ministral, Qwen).
\end{itemize}

\clearpage

\section{Supervised Post-Reason Tuning - Ablation Study: Training Corpus Dependency}
\label{app:dataset_ablation}

To isolate the source of the optimization benefits observed in our primary experiments, we conducted a data-centric ablation study. We held the Target-Conditioned Self-Distillation methodology constant but replaced the training corpus. We compared models fine-tuned on the Numina dataset (which is heavily biased toward complex, multi-step mathematical reasoning) against models fine-tuned on the Massive Multitask Language Understanding (MMLU) dataset (which is predominantly multiple-choice factual recall). 

Our objective was to determine whether the convergence advantages of our method are universally applicable or intrinsically tied to the structural depth of the training data.

\subsection{Optimization Dynamics: Reasoning vs. Factual Recall}

Target-conditioned self-distillation is designed to force the model to refine its intermediate chain of thought by masking the final answer. In mathematical problem-solving (Numina), these intermediate steps are critical algorithmic derivations. In factual tasks (MMLU), the intermediate steps are often shallow justifications that bottleneck on the base model's parametric memory.

\begin{figure}[htbp]
    \centering
    \includegraphics[width=0.75\textwidth]{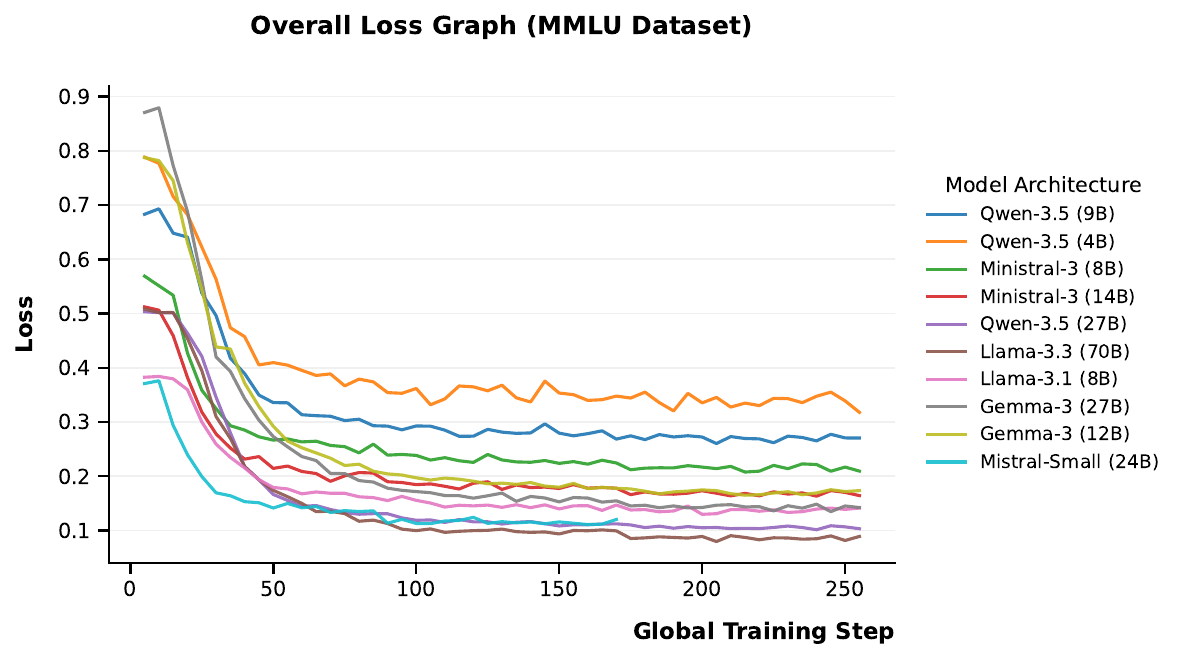} 
    \caption{Aggregate training loss convergence across all models on the MMLU dataset. When the training corpus is swapped from reasoning-heavy data (Numina) to factual data (MMLU), the optimization benefits of Target-Conditioned Self-Distillation are neutralized.}
    \label{fig:overall_mmlu_loss}
\end{figure}

This theoretical distinction is clearly visible in the training dynamics. When applied to the Numina dataset (as shown previously in Appendix \ref{app:extended_optimization}), our method achieves a significantly lower loss basin compared to standard distillation. However, when applied to the MMLU dataset (Figure \ref{fig:overall_mmlu_loss}), this advantage vanishes. The loss trajectories on the factual dataset remain entangled with baseline expectations, demonstrating that the masking mechanism requires complex reasoning pathways to optimize effectively.

\subsection{Downstream Benchmark Performance}

The neutralization of optimization benefits on the MMLU training set correlates directly with downstream performance. To quantify this, we evaluated the suite of MMLU-trained self-distilled models across our standard benchmark battery. 

\begin{table}[htbp]
    \centering
    \caption{Downstream performance comparison between MMLU SFT and Numina SFT. Values represent accuracy in percentage points. $\Delta$ is Numina SFT minus MMLU SFT, in accuracy points.}
    \label{tab:mmlu_vs_numina_benchmarks}
    \resizebox{\textwidth}{!}{%
    \begin{tabular}{lccccccccc}
        \toprule
        & \multicolumn{3}{c}{\textbf{BBH}} & \multicolumn{3}{c}{\textbf{GPQA}} & \multicolumn{3}{c}{\textbf{MMLU-Pro}} \\
        \cmidrule(lr){2-4} \cmidrule(lr){5-7} \cmidrule(lr){8-10}
        \textbf{Model} & MMLU & Numina & $\Delta$ & MMLU & Numina & $\Delta$ & MMLU & Numina & $\Delta$ \\
        \midrule
        \multicolumn{10}{l}{\textbf{Llama Family}} \\
        Llama-\(3.1\) (\(8\)B) & 41.28 & 43.68 & +2.40 & 32.81 & 34.16 & +1.35 & 35.67 & 36.80 & +1.13 \\
        Llama-\(3.3\) (\(70\)B) & 64.74 & 65.97 & +1.23 & 50.11 & 50.11 & 0.00 & 52.83 & 53.13 & +0.30 \\
        \midrule
        \multicolumn{10}{l}{\textbf{Gemma Family}} \\
        Gemma-\(3\) (\(12\)B) & 58.88 & 59.27 & +0.39 & 35.28 & 34.16 & -1.12 & 43.37 & 44.57 & +1.20 \\
        Gemma-\(3\) (\(27\)B) & 65.09 & 65.47 & +0.38 & 35.73 & 36.40 & +0.67 & 50.77 & 51.27 & +0.50 \\
        \midrule
        \multicolumn{10}{l}{\textbf{Mistral \& Ministral}} \\
        Ministral-\(3\) (\(8\)B) & 55.63 & 55.64 & +0.01 & 37.98 & 35.51 & -2.47 & 47.07 & 47.97 & +0.90 \\
        Ministral-\(3\) (\(14\)B) & 59.55 & 60.42 & +0.87 & 38.88 & 40.22 & +1.34 & 53.33 & 53.60 & +0.27 \\
        Mistral-Small (\(24\)B) & 59.55 & 59.94 & +0.39 & 42.47 & 45.17 & +2.70 & 54.13 & 54.53 & +0.40 \\
        \midrule
        \multicolumn{10}{l}{\textbf{Qwen-3.5 Family}} \\
        Qwen\(3.5\) (\(4\)B) & 53.16 & 53.99 & +0.83 & 43.60 & 43.37 & -0.23 & 45.23 & 45.67 & +0.44 \\
        Qwen\(3.5\) (\(9\)B) & 57.64 & 58.47 & +0.83 & 60.00 & 60.90 & +0.90 & 51.20 & 51.50 & +0.30 \\
        Qwen\(3.5\) (\(27\)B) & 68.24 & 68.78 & +0.54 & 61.57 & 63.60 & +2.03 & 65.50 & 65.53 & +0.03 \\
        \bottomrule
    \end{tabular}%
    }
\end{table}

As detailed in Table \ref{tab:mmlu_vs_numina_benchmarks}, utilizing the Numina training corpus yielded better performance compared to the MMLU corpus on BBH and MMLU-Pro for all 10 models, and on GPQA for 6 of 10 models. This ablation provides conclusive evidence that Target-Conditioned Self-Distillation thrives when applied to reasoning-dense datasets. The deep algorithmic trajectories inherent in Numina provide the necessary structural complexity to effectively guide the self-distillation objective and shape the loss landscape, proving the technique acts as a powerful reasoning multiplier rather than a simple knowledge injector.

\clearpage

\end{document}